\documentclass[sigconf]{aamas}

\usepackage{balance} 

\setcopyright{none}
\renewcommand\footnotetextcopyrightpermission[1]{}
\acmConference{Preprint}{2026}{}

\usepackage{amsthm,graphicx,paralist}
\usepackage[]{todonotes}
\usepackage{thm-restate}
\usepackage{hyperref}
\usepackage[ruled,noend,linesnumbered]{algorithm2e}
\usepackage{proof}
\usepackage{tikz}
\usepackage{forest}
\usepackage{subcaption}
\usetikzlibrary{shapes, arrows.meta, positioning}
\usepackage{float}
\usepackage{xcolor}

\SetCommentSty{mybluecomment}

\usepackage[capitalise]{cleveref}
\usepackage{tcolorbox}
\usepackage{multirow}
\usepackage{colortbl}
\usepackage{placeins} 

\newcommand{\distribution}{\mathsf{Dist}}

\newcommand{\mdp}{\mathcal{M}}
\newcommand{\mdpstates}{S}
\newcommand{\mdpinitstates}{s_{\mathsf{init}}}
\newcommand{\mdpactions}{A}
\newcommand{\mdpavailactions}{A}
\newcommand{\mdptransitions}{P}
\newcommand{\mdprewards}{R}
\newcommand{\mdppath}{\rho}
\newcommand{\mdppolicy}{\pi}

\newcommand{\last}{\mathsf{last}}

\newcommand{\pomdp}{\mathcal{P}}
\newcommand{\observationmap}{\mathcal{O}}
\newcommand{\observations}{Z}

\newcommand{\history}{h}
\newcommand{\histories}{\mathsf{Hist}_{\pomdp}}

\newcommand{\fsc}{\mathcal{F}}

\newcommand{\fscstates}{N}

\newcommand{\fscinitstate}{n_{\mathsf{init}}}
\newcommand{\fsctransitionmap}{\delta}
\newcommand{\fsclabelmap}{\gamma}

\newcommand{\storm}{\textsc{Storm}}
\newcommand{\prism}{\textsc{Prism}}
\newcommand{\paynt}{\textsc{Paynt}}

\DeclareGraphicsExtensions{.pdf, .png, .jpg}

\theoremstyle{plain}
\theoremstyle{definition}
\newtheorem{definition}{Definition}

\newtheorem{remark}{Remark}

\renewcommand{\Pr}{\mathbb{P}}
\newcommand{\Expectation}{\mathbb{E}}
\newcommand{\reward}{\mathsf{reward}}

\newcommand{\measure}[1]{\Pr_{#1}}

\newcommand{\DT}{\mathsf{DT}}
\newcommand{\dataset}{\mathsf{D}}

\newcommand{\entropy}{\mathsf{entr}}

\newcommand{\skipp}{\textsf{skip }}
\newcommand{\skippp}{\textsf{skip}}

\newcommand{\supp}{\mathsf{supp}}

\newcommand{\R}{\mathbb{R}}

\newcommand{\ObsDom}[1]{\mathsf{O}bs\mathsf{D}om_{#1}}

\SetKwProg{Fn}{Function}{:}{}

\newtheorem{example}{Example}

\title[Explainable Representation of Finite-Memory Policies]{Explainable Representation of Finite-Memory Policies\\ for POMDPs using Decision Trees}

\author{Muqsit Azeem}
\affiliation{
	\institution{Technical University of Munich}
	\city{Munich}
	\country{Germany}}
\email{muqsit.azeem@tum.de}

\author{Debraj Chakraborty}
\affiliation{
	\institution{Nanyang Technological University}
	\country{Singapore}}
\email{debraj.chakraborty@ntu.edu.sg}

\author{Sudeep Kanav}
\affiliation{
	\institution{Masaryk University}
	\city{Brno}
	\country{Czech Republic}}
\email{kanav@fi.muni.cz}

\author{Jan Kretinsky}
\affiliation{
	\institution{Masaryk University}
	\city{Brno}
	\country{Czech Republic}}
\email{jan.kretinsky@fi.muni.cz}

\begin{document}

\begin{abstract}
Partially Observable Markov Decision Processes (POMDPs) are a fundamental framework for decision-making under uncertainty but often require infinite memory, making implementation infeasible and many problems undecidable. While finite-memory policies provide a practical alternative, they remain complex and challenging to interpret.

To address this, we propose a novel \emph{representation} of finite-memory policies that is both (i) interpretable and (ii) smaller, enhancing explainability without sacrificing optimality. To that end, we combine Mealy machines and decision trees (DTs); the latter describing simple, stationary parts of the policies and the former describing how to switch among them.

We design a translation for finite-state-controller (FSC) policies from standard literature into our new representation, enhancing explainability and compactness while preserving optimality. Notably, our method seamlessly generalizes to other variants of finite-memory policies. Additionally, we identify unique properties of ``attractor-based'' policies, enabling the construction of even smaller, simpler representations. Finally, through multiple case studies, we illustrate the improved explainability and practicality of our approach.
\end{abstract}


\maketitle

\section{Introduction}

\paragraph*{POMDP and FSC}
\emph{Partially observable Markov decision processes (POMDPs)} provide a foundational framework for decision-making under uncertainty, where the agent lacks complete information about the current state. They have applications ranging from autonomous robotics and intelligent assistants~\cite{foka2007real} to healthcare systems~\cite{hauskrecht2000planning}.
Unfortunately, decision-making in POMDPs is inherently complex. In particular, infinite memory may be required for optimality for many objectives.
Not only are such policies impractical for real-world implementation, but this complexity also makes
finding the optimal policies particularly challenging and, in most settings, undecidable~\cite{madani2003undecidability}. 
Consequently, finite-memory policies are often considered, e.g.~\cite{andriushchenko2022, andriushchenko2023}.
Besides rendering various problems decidable (or at least allowing for efficient heuristics), they are clearly more amenable to implementation than general, infinite-memory policies.
Finite-memory policies are formalized as \emph{finite-state controllers (FSCs)}, essentially Mealy machines (finite-state automata with both input and output) where each state corresponds to a stationary (a.k.a.~memoryless) policy and the automaton transitions describe how to switch among them. 
However, while FSCs make POMDP policies implementable, certain issues arise concerning their explainability, which we discuss below and which motivate the need for more explainable representations of FSCs.

\paragraph*{Explainability.}
In many real-world domains---from autonomous driving to healthcare decision support---opaque or complex policies pose critical risks. Black-box policies can be hard to validate or trust, especially under uncertainty. Hence, explainable controllers are not merely preferable but also essential from the legal perspective~\cite{EUX21}.
The automata representation is often regarded as interpretable and even explainable \cite{bork2024learning} due to its graphical structure and typically low number of states. However, this issue is more complex as the automaton drawing does not depict the whole policy, notably:
(i)~Each state corresponds to a stationary policy, which is typically a \emph{table} with an action to be played for each observation; due to its often enormous size it is hard to explain and thus does not expose the goal of that particular policy.
(ii)~Each transition between states is typically labeled by many observations (often inconceivably many due to the combinatorial explosion), effectively hiding the reasons to switch to another policy.
While both issues prevent explainability, the latter one is more severe due to two reasons.
First, the size of the representation of transitions is even greater than that of the policies.
Second, while the former issue of representing a stationary policy has been tackled in literature, e.g.~\cite{dtcontrol2}, the latter issue of explainable transitions not so.
To summarize, the issues (i) and (ii) often obscure the decision logic: large action tables obscure the intent of each policy, and opaque transition labels conceal why the controller switches states. 
This lack of clarity undermines trust, especially when certification or debugging is required.

We address this lack of transparency systematically in several steps, exploiting the factored structure of states and observations.\\
$\bullet$ \emph{First}, we translate the tabular representations of the stationary policy in each FSC state into a decision tree (DT). 
While this has been done neither for POMDP, nor for randomizing policies, the generalizations are straightforward and thus a mere application of methods from the literature.\\
$\bullet$ \emph{Second}, while history-dependent policies have never been represented by DTs, we propose to do so by combining the FSC structure and the transition labels (being sets of observations) translated into DTs. 
While an alternative approach might be to encode the labels as Boolean formulae (over the observations or their variables), DT have the advantage of their hierarchical structure.
Indeed, while a formula (or its syntax tree) does not reflect the importance of each part, DT features its most important decisions close to the root with the less important ones near the leaves.
This allows for cheap and transparent pruning (ignoring the less important layers of the tree), offering a trade-off between precision and explainability.
In particular, in order to understand when stationary policies switch, one can easily understand the frequent simpler cases first and subsequently contemplate also the further corner-cases.

To summarize the two points, this translation from FSC to ``DT-FSC'' replaces opaque action and transition tables with interpretable decision trees that expose how observations guide both action selection and memory updates.
While making the representation more concise (and interpretable due to the factored structure), we retain the exact functional equivalence to the original policy.
On top of that, searching the tables is replaced by evaluating the decision trees, which are isomorphic to if-then-else code, and can thus be executed orders of magnitude \emph{faster} \cite{kiesbye2022planning}.\\
$\bullet$ \emph{Third}, we profit from the new DT-FSC representation, in order to provide an \emph{explanation} of the policy.
On the one hand, prior work on explainable policies via DT~\cite{dtcontrol, dtcontrol2} adopts the widely used view that \emph{the size of the DT} is a practical proxy for explainability: smaller trees are easier to inspect, understand, and trust.
This perspective is further supported by recent studies in human-robot interaction and decision-making tasks, e.g.~\cite{verma2025interpretability}.
Consequently, we adopt this proxy of explainability for the individual trees.
On the other hand, we have to critically assess this perspective when it comes to history-dependent policies since the sequential, dynamic changes in behaviour pose an inherent additional challenge for explainability.
We scrutinize the resulting structure in the light of several case studies.
To that end, we cast away quantitative proxies of explainability and observe how our result stands the ultimate (however subjective) test of reaching the ``aha, now I see''-moment.
We have observed the following:
(i) In contrast to the large FSC, from the DT-FSC representation one can trivially read off several English sentences describing completely the rules of the policy.
(ii) This in turn does \emph{not} itself constitute an intuitive explanation of the policy behaviour to be understood by a human, e.g. a colleague,
mostly due to the sequential composition of varying behaviours.
(iii) An intuitive explanation can be derived in each case, using manual effort.
For instance, the rules ``in the initial configurations we move to the south'' and``the direction at some point changes to the east, then to the north, then to the west'' must be manually turned into ``we are going in circles''.
(iv) Most importantly, while the DT-FSC representation allowed for the manually gained insight, the original tabular FSC with its sheer size and non-factored, structureless information did not.
We conclude that our new DT-FSC can serve as the first technical step towards generating actual explanations of history-dependent policies.

\noindent\textbf{Our technical contribution} can be summarized as follows:
In particular, we propose a new data structure to represent FSCs (history-dependent policies), merging the advantages of the Mealy machines and the interpretable \emph{decision trees} (DTs). 
The latter is used not only for representing the stationary policies similarly to literature, e.g.~\cite{dtcontrol2}, but importantly also for the automata transitions.
This reduces the redundancy by grouping related observations, which together with the DT structure yields a panoramic view of decision paths of conceivable size, and highlights critical variables and decision points.
Besides, we identify specific properties of ``attractor-based'' policies \cite{JungesJS20}, which allow us to construct yet simpler and smaller representations with more compact and fewer transitions.
Notably, \emph{our transformation procedure alters only the representation obtained from existing tools, preserving both the policies and their optimality.}
Further, as it is \emph{modularly} defined on parts of the original policy, it captures the parts separately, simplifying the explanation compared to a larger monolith.
As a result, lowering the barrier for their human inspection, validation, and debugging moves us closer to making formally verified decision-making systems both transparent and usable in practice.

\subsubsection*{Related Work}
\textbf{POMDPs} can be solved using various
exact~\cite{smallwood1973,geffner1998solving} or simulation-based~\cite{silver2010monte} techniques.
Formal verification tools like \storm~\cite{storm} use \emph{belief}-based
analysis~\cite{aastrom1965optimal} to find an optimal policy.  
For qualitative objectives such as almost-sure reachability, SAT/SMT-based methods (both one-shot~\cite{almost-sure-POMDP-Kirsh2016} and iterative approach~\cite{JungesJS20}) are shown to be effective.
We build on these approaches by transforming their FSC outputs into more interpretable forms. Our technique can take specific advantage of this iterative SMT-based approach.

\smallskip
\noindent\textbf{FSCs} representing POMDP policies have been synthesized by inductive synthesis approaches of \paynt~\cite{paynt}, learnt by using $L^*$~\cite{bork2024learning}, 
extracted from recurrent policy networks~\cite{rnn-2-fsc-Koul2018,rnn-2-fsc-task-aware}
or computed using a combination of belief-based and inductive approaches together~\cite{andriushchenko2023}.
Our technique takes these types of controllers as input and generates their more explainable representation.
Another relevant model is the \emph{plan graph} \cite{KAELBLING199899}: a Moore machine FSC where the nodes are labeled by actions and edges are labeled by observations. In our Mealy-machine FSC, the observation-action mapping is explicit on each edge, making it easier to read and audit ``if observe $o$, then take  $a$ and go to the next stationary policy'' without mentally combining node and edge labels. Further, Mealy controllers are generally smaller, e.g., a policy can be a one‑node Mealy controller but generally needs $|A|$ nodes as a policy graph.

\smallskip
\noindent\textbf{Explainability in Decision Making.}
Explainability in sequential decision making 
has been explored through various lenses in the AI community.
Prior works include policy summarization~\cite{khan2009minimal, hayes2017transparency}, counterfactual reasoning~\cite{madumal2020explainable}, and causal models~\cite{pearl2009causality}, which seek to make policy decisions understandable through simplification or causal explanation.
More generally, explainability is often defined as the degree to which a human can understand the cause of a decision~\cite{miller2019explanation}, or trace the steps taken by an agent~\cite{guidotti2018survey} in a black-box setting.
In our work, we (i) adopt the framework suggesting small decision trees as means to explain a controller at one glance (rather than a single decision) \cite{DBLP:conf/cav/BrazdilCCFK15,dtcontrol,dtcontrol2,kiesbye2022planning,budde2024digging,andriushchenko2025small}, and
(ii) sanity-check its compatibility with explanations being a few English sentences triggering the aha moment. 
(For the sake of readability, we do not drag the reader into the details of cognitive complexity \cite{campbell2017cognitive} to define the latter more formally.)

\smallskip
\noindent\textbf{Decision Trees (DTs)} provide a natural mechanism for achieving explainability in this setting.  
Their hierarchical, rule-based structure aligns closely with human reasoning and supports transparent analysis of decision logic~\cite{dtcontrol,dtcontrol2}.  
DTs have been widely used in data mining~\cite{maimon2014data}, supervised learning~\cite{breiman2017classification}, and controller synthesis, including the representation of controllers in 
different systems \cite{counterexample-learning,DBLP:conf/ijcai/BoutilierDG95,neider-fmcad19}.  
The dtcontrol tool~\cite{dtcontrol} has demonstrated that stationary policies can be compactly and effectively represented using DTs.  
Building on this, our work proposes the use of DTs beyond stationary policies to finite-memory controllers through a modular transformation that embeds interpretability directly into the policy representation.
By representing both action selection and memory transitions as decision trees, our approach produces a globally explainable controller without compromising correctness or optimality.

Technically similar structures, such as decision diagrams (BDDs), have also been used to represent transitions and output in symbolic automata~\cite{fisman2023inferring}. However, despite being more compact in some settings, they have a few disadvantages in this context.
In particular, their interpretability is hindered by non-Boolean variables being converted to bitvectors using bit-blasting; they lack variability of ordering in different branches, which is important for controllers; they handle don't care inputs in less efficient ways than DT.
In contrast, more interpretable DT are easier for humans to follow,  and are compatible with tools like dtControl~\cite{dtcontrol2}, which support expert-in-the-loop pruning and analysis.
Due to these and further reasons they have not been used as much as DT \cite{dtcontrol}, although the recent work \cite{chakraborty2025explaining} diminishes the gap.

\section{Preliminaries}
A \emph{probability distribution} on a countable set $S$ is a function $d:S\to [0,1]$ such that $\sum_{s\in S} d(S)=1$.
\emph{Support} of a probability distribution $d$, is the set 
	$\supp(d)=\{s\in S\mid d(s)>0\}$.
We denote the set of all probability distributions on the set $S$ as $\distribution(S)$. 

\begin{definition}[POMDP]
	A \emph{partially observable MDP (POMDP)} is a tuple $\pomdp =(\mdpstates, \mdpactions, \mdptransitions, \mdprewards, \mdpinitstates, \observations, \observationmap)$ where $\mdpstates$ is a countable set of states, $\mdpactions$ is a finite set of actions, 
	$\mdptransitions: \mdpstates\times \mdpactions\rightharpoonup \distribution(\mdpstates)$ is a partial transition function,
	$\mdprewards: \mdpstates\times \mdpactions\to \R$ is a reward function,
	and $\mdpinitstates\in\mdpstates$ is the initial state, $\observations$ is a finite set of observations, and $\observationmap:\mdpstates\rightarrow\observations$ is an observation function\footnote{Deterministic observation functions are sufficient as the probabilistic aspect of the observation function can be encoded in the transition function \cite{chatterjee2016optimal}. 
		Thus our definition with a deterministic observation function does not lose generalization, 
		but rather, simplifies the notation and also make it more explainable.} that maps each state to an observation.	
\end{definition}

	We assume that the observation space is factored, i.e. $ \observations = \prod_{i=1}^{k} \observations_i$ where each component $ \observations_i$ represents a distinct feature of the observation, meaning any observation $z\in \observations$ can be represented as a tuple of features  $(z_1, z_2, \dots, z_k)$, where each $ z_i \in \observations_i $. 
	Note that, this assumption is quite natural. 
	In many real-world scenarios, observations come from a set of independent components, each reflecting a specific aspect of the state. 
	For example, in robotics, the observation can include the robot's position, battery level, and different sensor readings. 

	We define the set of \emph{enabled actions} in $s\in \mdpstates$ by $\mdpavailactions(s) = \{a\in \mdpactions \mid \mdptransitions(s,a)\textrm{ is defined}\}$.
	For a POMDP $\pomdp$, a \emph{path} is
	a sequence 
	$s_0a_0s_1a_1s_2\ldots$ such that for all $i$,
	$a_i\in \mdpavailactions(s_i)$ and $\mdptransitions(s_i,a_i)(s_{i+1})>0$.  
	We extend the notion of observations to the paths in a natural way:
        for a path $\rho=s_0 a_0 s_1 a_1 \cdots$,
		$\observationmap(\rho)$ is the sequence 
		$\observationmap(s_0) a_0 \observationmap(s_1) a_1 \cdots$.
		A \emph{history} is a finite sequence of observations and actions
		$\history = o_0 a_0 \ldots a_{i-1} o_i$ such that there exists a finite path
		$\mdppath = s_0 a_0 s_1 \ldots s_i$ with $o_j = \observationmap(s_j)$ for $0 \leq j \leq i$.
		We denote the set of all histories of $\pomdp$ by $\histories$.

\begin{definition}[Policy for POMDPs]
	A \emph{policy} for a POMDP $\pomdp$ is a function $\mdppolicy : \histories\to\distribution(\mdpactions)$ mapping histories to actions.
\end{definition}

A policy $\mdppolicy$ is \emph{deterministic} if $|\supp(\mdppolicy(\history))| = 1$ for all paths $\history\in \histories$. Otherwise, it is \emph{randomized}. 
A policy $\mdppolicy$ is called \emph{stationary} or \emph{memoryless} if it depends only on the last observation.

Policies resolve nondeterminism by converting a POMDP to a fully observable Markov chain. 
This Markov chain induces a unique probability measure $\measure{\mdppolicy,s_0}$ over paths starting in initial state $s_0$~\cite{BaierKatoen}.

\paragraph{Policy Synthesis.}
The \emph{policy synthesis} problem consists of finding a policy that satisfies a certain objective $\varphi$. 
In this paper, we consider infinite-horizon reachability or expected total reward objectives. 
Given a set of target states $T\subseteq S$, we denote $\measure{\mdppolicy,s_0}(\Diamond T)$
to be the probability of reaching a state in $T$ following the policy $\mdppolicy$.
In the case of \emph{almost-sure} reachability objective, a policy $\mdppolicy$ is \emph{winning} from $s_0$ if $\measure{\mdppolicy,s_0}(\Diamond T)=1$.
In this case, we denote $\Expectation_{\mdppolicy,s_0}(\reward(\Diamond T))$
to be the expected total reward accumulated till we reach the states in $T$.
Alternatively, we can look at \emph{quantitative} objectives:
given an objective $\varphi$ and a threshold $\lambda\in(0,1)$, a policy is winning 
if $\measure{\mdppolicy,s_0}(\Diamond T)\geq\lambda$.

\paragraph{Finite-State Controller (FSC)} We provide an abstract definition of FSC and then give instantiations used in literature.
\begin{definition}[FSC]\label{def:2-obs-fsc}
	A finite state controller (FSC) is a tuple $\fsc=(\fscstates,\fscinitstate, \fsctransitionmap, \fsclabelmap)$ where 
	$\fscstates$ is a finite set of nodes,
	$\fscinitstate\in\fscstates$ is the initial node,
	$\fsctransitionmap\colon\fscstates\times\ObsDom{\fsctransitionmap}\to\fscstates$ is the transition function,
	$\fsclabelmap\colon\fscstates\times\ObsDom{\fsclabelmap}\to\mdpactions$ is the action mapping,
	where $\ObsDom{\fsctransitionmap}$ and $\ObsDom{\fsclabelmap}$ are observation domains over $\observations$.
\end{definition}

Typically, in an FSC, $\ObsDom{\fsclabelmap}= \observations$ and $\ObsDom{\fsctransitionmap} = \observations$.
Different works in the literature adopt slightly different notions of these observation domains. For instance, some definitions let transitions depend only on the current observation~\cite{bork2024learning}, i.e., $\ObsDom{\fsctransitionmap} = \observations$, while others~\cite{JungesJS20,andriushchenko2023} use both the current and the next (posterior) observation, i.e., $\ObsDom{\fsctransitionmap} = \observations \times \observations$.
In this case, in state $s$ with observation $z$, an agent following an FSC $\fsc$ executes the action $a = \fsclabelmap(n, z)$ and then based on the current and the next observation $z$ and $z'=\observationmap(s')$, the FSC evolves to node $n' = \fsctransitionmap(n,z,z')$.

\begin{remark}
	Our formulation abstracts over these choices by introducing the generic domains $\ObsDom{\fsclabelmap}$ and $\ObsDom{\fsctransitionmap}$.
	This unifies the variants used in the literature~\cite{JungesJS20,andriushchenko2023,bork2024learning}, and our framework can handle controllers defined under any such convention.
\end{remark}

\begin{definition}[Decision Tree (DT)]
	Given an input space $\mathbb{X}$ and an output space $\mathbb{A}$, a \emph{decision tree} $T: \mathbb{X} \rightarrow \mathbb{A}$ is a binary tree where each inner node  $v$ is labeled with a \emph{predicate} $P_v: \mathbb{X} \to \{\texttt{true}, \texttt{false}\} $. 
	Each edge corresponds to an outcome of the predicate: solid for \texttt{true}, dotted for \texttt{false}. 
	Each leaf $\ell$ is labeled with an output $a_\ell \in \mathbb{A}$. 
	The output $T(x)$ for an input $x \in \mathbb{X}$ is determined by evaluating predicates from the root to a leaf,  and returning the output label $a_\ell$ of that leaf.
\end{definition}

\paragraph{Decision Trees for Policy Representation.}~DT learning algorithms can exploit the structure of factored observations to represent policies in a compact and explainable way.
A stationary policy can be exactly represented by a DT~\cite{dtcontrol}, where the input is the state set $\mdpstates$ and the output labels correspond to actions $\mdpactions$ suggested by the policy. 
Unlike conventional DTs in machine learning, which aim to generalize and may tolerate classification errors, DTs in this context must represent the policy exactly to ensure correctness. The following sections extend this idea to non-stationary policies.

\section{An Illustrative Example Explaining Policy Representation}
\label{sec:solution}
We illustrate how a non-stationary POMDP policy can be represented as an FSC. 
As a running example, we use the cheese-maze model~\cite{littman1995learning} as detailed in~\cref{ex:maze}.

\begin{figure}[t]
	\centering
	\includegraphics[width=0.25\textwidth]{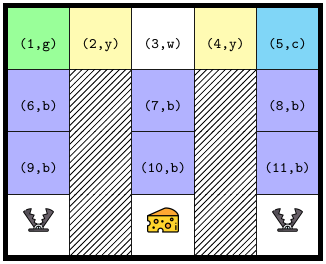}
	\caption{Maze with state-observation labels; colors indicate observations: green (g), yellow (y), white (w), cyan (c), and blue (b). Shaded cells are inaccessible.}
\label{fig:maze}
\end{figure}

\begin{example}[Maze]\label{ex:maze}
	Consider the maze in \cref{fig:maze}, modeled as a POMDP. 
	A mouse is initially placed randomly in the grid in a state with observation blue.
	Its objective is to reach the target (cheese) while avoiding the traps. Each cell represents a distinct state, and the mouse can move in one of the four possible cardinal directions at each step unless blocked by a wall. 
	However, the mouse's observations are limited to local wall configurations, making it unable to distinguish between cells with identical surroundings (indicated by the same color).
	Consequently, there are only seven distinct observations, with the target and traps considered as separate ones.
	
\end{example}

\paragraph{Finite State Controller Representation.}
In \cref{ex:maze}, the observation space is structured with the following boolean variables: \texttt{CanGoDown}, \texttt{CanGoLeft}, \texttt{CanGoRight}, \texttt{CanGoUp}, which indicate if the mouse can move in each direction. 
Additionally, there are three more observable variables: \texttt{bad} to denote whether the mouse is in a trap, \texttt{clk} to denote if the mouse has entered the maze and \texttt{goal}, which indicates whether the mouse is at the target. \cref{fig:mealy-standard-table-maze} shows an FSC representation of a winning policy. In the tables, each observation is represented as a vector over these seven variables.

\begin{figure}[t]
	\centering
	\colorbox{blue!10}{ 
		\resizebox{\columnwidth}{!}{
			\begin{tikzpicture}[node distance=7cm, auto, thick, >=Stealth]
				\tikzstyle{state} = [rectangle, draw, rounded corners, minimum width=4cm, minimum height=1.5cm, text width=3.8cm, align=center, scale=0.8, fill=white!5]
				
				\node[state] (n2) {\textbf{$n_0$} \\[0.2cm]
					\begin{tabular}{c|c}
						Observation & Action \\ \hline
						$0,1,1,0,0,1,0$ & left \\
						$0,0,0,0,0,0,0$ & INIT \\
						$1,1,0,0,0,1,0$ & left \\
						$1,0,0,1,0,1,0$ & up \\
					\end{tabular}
				};
				
				\node[state, right of=n2] (n1) {\textbf{$n_1$} \\[0.2cm]
					\begin{tabular}{c|c}
						Observation & Action \\ \hline
						$1,0,1,0,0,1,0$ & right \\
						$1,0,0,1,0,1,0$ & down \\
						$1,1,1,0,0,1,0$ & down \\
						$0,1,1,0,0,1,0$ & right \\
					\end{tabular}
				};
				
				\node[left=0.7cm of n2] (start) {};
				\draw[->, thick] (start) to[out=-20, in=180] (n2);
				
				\node[rectangle, draw, minimum size=0.5cm] (squarebox1) at ($(n2)!0.5!(n1)$) {$t_0$};
				
				\draw[-, thick] (n2) to (squarebox1.west);
				\draw[->, thick] (squarebox1)  to [out=0, in=160] (n1.west);
				
				\node[rectangle, draw, minimum size=0.5cm] (squarebox2) at ($(n1.north) + (0,.65)$) {$t_1$};
				
				\draw[-, thick] (n1.110) to[out=140, in=180] (squarebox2.west);
				\draw[->, thick] (squarebox2.east) to[out=0, in=40] (n1.70);
				
				\draw[->, thick] (squarebox1) edge[loop, in=80, out=110, looseness=1] node {} (n2);	
			\end{tikzpicture}
		}
	}
	
	\begin{minipage}{\textwidth}
		\begin{minipage}{0.26\linewidth}
			\scriptsize
			\colorbox{green!10}{ 
				\begin{tabular}{c|c}
					Next observation & Next node \\ \hline
					$0,1,1,0,0,1,0$ & $n_0$\\
					$1,1,0,0,0,1,0$ & $n_0$\\
					$1,1,1,0,0,1,0$ & $n_1$\\
					$1,0,0,1,0,1,0$ & $n_1$\\
					$0,0,0,0,0,0,0$ & $n_0$\\
					$1,0,1,0,0,1,0$ & $n_1$\\
				\end{tabular}
			}
		\end{minipage}%
		\hspace{0.25cm}
		\begin{minipage}{0.42\textwidth}
			\scriptsize
			\colorbox{green!10}{ 
				\begin{tabular}{c|c}
					Next observation & Next node \\ \hline
					$0,1,1,0,0,1,0$ & $n_1$\\
					$1,1,1,0,0,1,0$ & $n_1$\\
					$1,0,0,1,0,1,0$ & $n_1$\\
					$1,0,1,0,0,1,0$ & $n_1$\\
					&\\
					&\\
				\end{tabular}
			}
		\end{minipage}
	\end{minipage}
	
	\caption{An FSC represented with explicit tables. Each observation is encoded as a vector over the seven observale variables. The transition tables $t_0$ and $t_1$ are below nodes $n_0$ and $n_1$, respectively. In this example, only the posterior
		observations are relevant to make decisions, so only the posterior observations ($z'$) are shown in the tables.}
	\label{fig:mealy-standard-table-maze}
\end{figure}

For each node of the FSC, there are two associated tables: (i) an \textit{action table} that specifies which action to take for each observation, and (ii) a \textit{transition table} that determines the next node based on the current and next observation.
At each step, the mouse observes its environment and selects an action according to the action table of the current node.  
After executing the action and receiving a new observation, the mouse may transition to a new controller node (e.g., from $n_0$ to $n_1$) based on the transition table.
Each node thus defines a stationary policy that the mouse follows while its controller remains in that node.
\emph{This mechanism allows the FSC to encode finite memory: the current node implicitly tracks part of the observation history.}

\begin{figure}[t]
	\centering
	\colorbox{blue!10}{ 
		\resizebox{\columnwidth}{!}{ 
			\begin{tikzpicture}[node distance=5.5cm, auto, thick, >=Stealth]
				
				\tikzstyle{state} = [rectangle, draw, rounded corners, minimum width=3cm, minimum height=5cm, align=center, scale=0.7, fill=white!10]
				
				\node[state] (n2) {\textbf{$n_0$} \\[0.2cm]
					\colorbox{white!10}{\includegraphics[width=3.5cm]{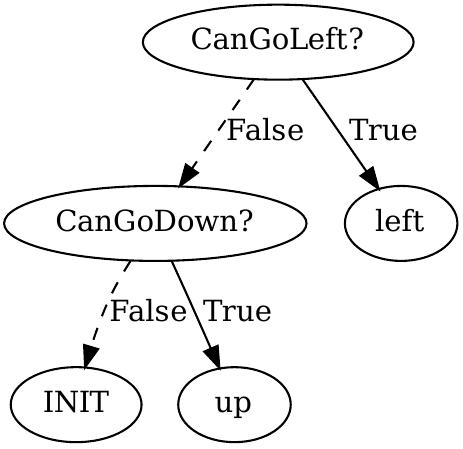}}
				};
				\node[state, right of=n2] (n1) {\textbf{$n_1$} \\[0.2cm]
					\colorbox{white!10}{\includegraphics[width=3.5cm]{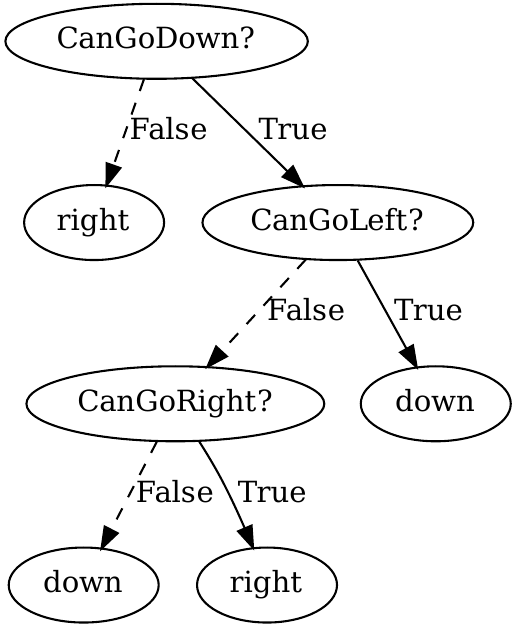}}
				};
				
				\node[above left=0.5cm of n2] (start) {};
				\draw[->, thick] (start) -- (n2);
				
				\draw[->, thick] (n1) edge[loop above, in=80, out=110, looseness=1] node {} (n1);
				\draw[->, thick] (n2) edge[bend left] node {} (n1);
				\draw[->, thick] (n2) edge[loop above, in=80, out=110, looseness=1] node {} (n2);
			\end{tikzpicture}			
		}
	}
	\colorbox{green!10}{ 
		\resizebox{\columnwidth}{!}{ 
			\begin{tikzpicture}		
				\node[below=0.2cm of n2] (transition_dt2) {
					\colorbox{white!10}{\includegraphics[width=2.5cm]{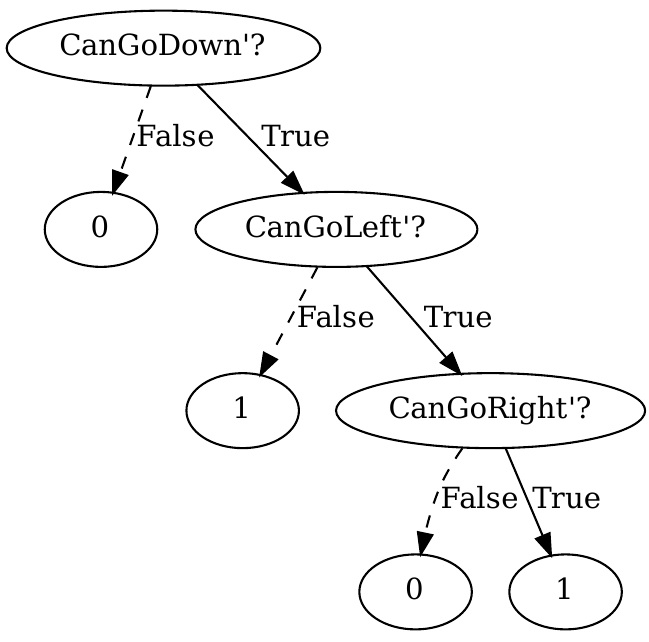}}
				};
				
				\node[below=0.2cm of n1] (transition_dt1) {
					\colorbox{white!10}{\includegraphics[width=0.7cm]{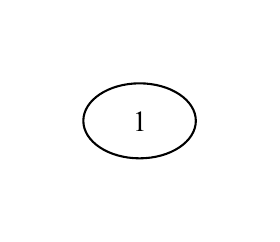}}
				};
			\end{tikzpicture}			
		}
	}
	\caption{{Explaining FSCs using DTs. For an observation variable $z$, $z'$ would denote the next observation, \textit{i.e.}, the observation the mouse sees after taking the action. A DT in a node describes a positional policy, a DT below a node describes the next node for the next observation (curved arrows in the FSC indicate possible transitions). Here, the solid edges in the DT are used to denote \texttt{true}, dotted to denote \texttt{false}.}}
	\label{fig:dt-fsc-maze}
\end{figure}

\paragraph{Decision-Tree-Based Explainability.}
Instead of using large tables to specify the stationary policy and transitions at each node of FSC, we translate both into decision trees, resulting in a DT-FSC (see~\cref{fig:dt-fsc-maze}). 
The DT-FSC in~\cref{fig:dt-fsc-maze} has two nodes, each with one DT for action selection and one for transition between the FSC nodes.
While the FSC nodes still capture history, the DTs exploit the structure in observations to make the representation more compact and interpretable.
For instance, the predicate \texttt{CanGoUp} is absent from all DTs, showing it plays no role in decision making. Thus, complementing FSCs with DTs provides a clearer and more structured explanation than policy tables.

In this example, the mouse starts with initial policy in node~$n_0$. If it is not yet placed in the grid, it first plays \texttt{INIT}, which randomly places it in a state with observation blue.
From there, it moves \texttt{up} on blue and \texttt{left} on yellow or cyan. It remains in node~$n_0$, effectively waiting until the next (posterior) observation becomes green or white---at which point it switches to the policy in node~$n_1$. 
The policy in~$n_1$ directs the mouse to go \texttt{right} when it sees green or yellow, continuing until it reaches white.
On white or blue, it then suggests moving \texttt{down} to eventually reach the cell containing the cheese (see~\cref{fig:maze-policy} in~\cref{app:illustrative_example} for a pictorial representation).
Note that conflicting behaviors for observations like blue and yellow are resolved through memory: the FSC node captures this context.

For smaller grid-based examples, such as this one, a pictorial policy representation may be easier to understand.
However, the DTs provides a general, domain-independent representation.
Moreover, while the grid policy size grows linearly with the problem, the DT-based representation may remain compact if the decision structure stays simple.

\section{Explainable Representation of FSCs via Amalgamation with DTs}

In this section, we formally describe how explainability of FSCs can be enhanced using DTs by introducing a specific data structure we call \emph{DT-FSC}. 
We then discuss its general applicability and further describe a particular scenario where FSCs (and their corresponding DT-FSCs) can be built using an iterative approach, which simplifies the explanation process. 
For this scenario, we also introduce a specialized data structure that offers the same level of explainability in a more compact form.

\begin{definition}[DT-FSC]
	Given an FSC $\fsc=(\fscstates,\fscinitstate, \fsctransitionmap, \fsclabelmap)$, we define its DT-FSC representation as follows:
	$$\fsc_{\DT}=(\fscstates,\fscinitstate,\{\DT_{\fsctransitionmap,n}\}_{n\in \fscstates},\{\DT_{\fsclabelmap,n}\}_{n\in \fscstates})\,,$$ where, 
	for each $n\in\fscstates$,
	$\DT_{\fsctransitionmap,n}$ is the decision tree representation of $\fsctransitionmap_n$,
	and $\DT_{\fsclabelmap,n}$ is the decision tree representation of $\fsclabelmap_n$.
\end{definition}

We define the semantics of a DT-FSC via a single-step execution, analogous to the single-step execution of an FSC, where $\fsctransitionmap$ and $\fsclabelmap$ are replaced by ${\DT_{\fsctransitionmap,n}}$ and ${\DT_{\fsclabelmap,n}}$, respectively.%
\cref{fig:dt-fsc-maze} shows the DT-FSC representation of the policy for the FSC in \cref{fig:mealy-standard-table-maze}. 
The trees inside the nodes describe the stationary policies. The trees below the node describe the transition between the nodes.
Although the transition function formally depends on both current and next observations, the learned decision trees simplify this by ignoring irrelevant inputs.
For example, the transition DT in \cref{fig:dt-fsc-maze} relies only on the posterior observation, even though the original transition function considers both.

\subsection{From FSC to DT-FSC: Translation Algorithm} 
\cref{alg:dt_fsc_creation} outlines the translation of an FSC into a DT-FSC in two phases: (1)~dataset construction and (2)~decision tree learning.

\begin{algorithm}[t]
	\caption{Translation from FSC to DT-FSC}
	\label{alg:dt_fsc_creation}
	\KwIn{FSC $\fsc = (\fscstates,\fscinitstate, \fsctransitionmap, \fsclabelmap)$}
	\KwOut{DT-FSC \scalebox{0.91} {$\fsc_{\DT} = (\fscstates, \fscinitstate, \{\DT_{\fsctransitionmap,n}\}_{n \in \fscstates}, \{\DT_{\fsclabelmap,n}\}_{n \in \fscstates})$}}
	
	\For{$n \in \fscstates$}{
		\tcp{Phase 1: Dataset construction}
		Construct dataset $\dataset_{\fsclabelmap,n}$ from $\fsclabelmap$\;
		Construct dataset $\dataset_{\fsctransitionmap,n}$ from $\fsctransitionmap$\;
		
		\tcp{Phase 2: Decision tree learning}
		$\DT_{\fsclabelmap,n} \gets \texttt{LearnDT}(\dataset_{\fsclabelmap,n})$\;
		$\DT_{\fsctransitionmap,n} \gets \texttt{LearnDT}(\dataset_{\fsctransitionmap,n})$\;
	}
	\Return $\fsc_{\DT}=(N, n_{\text{init}}, \{\text{DT}_{\delta,n}\}_{n \in N}, \{\text{DT}_{\gamma,n}\}_{n \in N})$\;
	\hfill \raisebox{0.30ex}{\rule{0.25\linewidth}{0.4pt}}~
	Helper: \texttt{LearnDT}~
	\raisebox{0.30ex}{\rule{0.25\linewidth}{0.4pt}} \hfill

	\Fn{\texttt{LearnDT} $(\dataset)$}{
		\If{all $(x, y) \in \dataset$ have the same label $y$}{
			\Return $\texttt{Leaf}(y)$\;
		}
		\Else{
		$p \gets \texttt{FindBestPredicate}(\dataset)$\; \label{ln:find-best-split}
		$\dataset_{\mathit{left}} \gets \{(x, y) \in \dataset \mid p(x)\}$\;
		$\dataset_{\mathit{right}} \gets \{(x, y) \in \dataset \mid \neg p(x)\}$\;
		\Return \scalebox{0.97} {$\texttt{Node}(p, \texttt{LearnDT}(\dataset_{\mathit{left}}), \texttt{LearnDT}(\dataset_{\mathit{right}}))$}\;
	}
	}
\end{algorithm}

\paragraph{(1)~Dataset Construction.}
For each node $n\in\fscstates$ in the FSC $\fsc=(\fscstates,\fscinitstate, \fsctransitionmap, \fsclabelmap)$, we extract two functions:
i) $\fsclabelmap_n\colon \ObsDom{\fsclabelmap}\to\mdpactions$ 
where $\fsclabelmap_n(o_{\fsclabelmap})=\fsclabelmap(n,o_{\fsclabelmap})$, and 
ii) $\fsctransitionmap_n\colon \ObsDom{\fsctransitionmap}\to\mdpactions$ 
where $\fsctransitionmap_n(o_{\fsctransitionmap})=\fsctransitionmap(n,o_{\fsctransitionmap})$. 
A function $f\colon X\to Y$ with a finite domain $X$ can be treated as a finite set $\dataset \subseteq X\times Y$, a dataset of feature-label pairs.
We obtain $\dataset_{\fsclabelmap,n}$  from $\fsclabelmap_n$ and $\dataset_{\fsctransitionmap,n}$ from $\fsctransitionmap_n$.

\paragraph{(2)~Decision Tree Learning.}
For $n\in\fscstates$, we learn the decision trees $\DT_{\fsclabelmap,n}$ and $\DT_{\fsctransitionmap,n}$ representing the mapping $\fsclabelmap_n$ and $\fsctransitionmap_n$ exactly.
A DT learning algorithm recursively splits the dataset to minimize the entropy of the resulting subsets.
The entropy of a dataset $\dataset$ over labels $Y$ is defined as
$\entropy(\dataset)=-\sum_{y\in Y}p_y\log(p_y)$
where $p_y$ is the fraction of examples labeled $y$, i.e.\  $p_y=\frac{|(x,y)\in \dataset|}{|\dataset|}$.
The best split is chosen via \texttt{FindBestPredicate} (line~\ref{ln:find-best-split} of \texttt{LearnDT} helper function) as the one that minimizes the total entropy after the split:
$\entropy(\dataset_{\mathit{left}}) + \entropy(\dataset_{\mathit{right}})$.
Since our goal is exact representation, we recurse until all examples in a subset share the same label.
The resulting DT-FSC $\fsc_{\DT}$ preserves the original FSC behavior exactly while offering a more interpretable representation.

\subsubsection{General Applicability of our DT-FSCs}
While optimal policies for POMDPs may require infinite memory in general~\cite{madani2003undecidability}, in practice, ``good enough'' finite-memory policies are considered to balance expressiveness with tractability~\cite{andriushchenko2023,bork2024learning}. 
For instance, \cite{andriushchenko2023} uses an anytime inductive synthesis approach that iteratively searches for optimal FSCs of increasing sizes. \cite{bork2024learning} uses \storm~to find a policy and then learns an FSC from it. 
Our DT-FSC translation framework is broadly applicable to such settings.

\storm~uses partial exploration~\cite{bork2020} and produces a randomized policy for the unexplored regions, referred to as \emph{cutoffs}.
Even in such cases, we can construct a DT-FSC that explains both the stationary policy and the transition function for each node.
Since the number of cutoffs is finite, the policy suggests only finitely many randomized actions---allowing us to build  DTs with a finite number of leaf labels.

\begin{remark}
	\emph{The performance of our DT-FSC policy remains the same as the one of the input policy. In particular, the action and transition functions are total, and the DTs are trained to overfit the data, guaranteeing functional equivalence with the input FSC, as done in~\cite{dtcontrol}.}
\end{remark}

\subsection{\skippp-FSC Optimization for Attractor-based Policies (for Almost-sure Reachability)}

For almost-sure reachability objectives, we construct a policy to reach the target by iteratively computing the attractor\footnote{An attractor is the set of observations from which a policy can ensure reaching the target region, possibly in multiple steps.} of the winning region (target set), following the approach of~\cite{JungesJS20}. For such policies, we provide an even more compact representation.
Note that, here we have $\ObsDom{\fsclabelmap} = \observations$ and $\ObsDom{\fsctransitionmap} = \observations\times \observations$.
Starting from target states, in each iteration, an observation-based stationary policy is obtained 
that either reaches the target or reaches an observation from which we have already found a winning policy in a previous iteration. 
In the second case, the policy would switch to the previously calculated winning policy.
This iterative procedure also helps us construct the FSC representing a winning policy.
At iteration $i$, we add a new node $n_i$ to the FSC.
The policy $\sigma_i$ found in the iteration would describe the action mapping, 
\textit{i.e.}, $\fsclabelmap(n_i,z)=\sigma_i(z)$.
If the policy requires switching to a previously computed policy, this is captured by the transition function $\fsctransitionmap$. 
This can be conceptualized as follows: 

	\emph{Case 1:} Based on the current observation $z$, the policy can suggest an action $a$ and does not switch to any other policy. In that case, in the constructed FSC, $\fsclabelmap(n_i,z)=a$ and the node of the FSC would not change. 
	
	\emph{Case 2:} Upon encountering an observation $z'$, the policy can switch to a previously computed policy $\sigma_j$ (without taking any action).
	In that case, in the constructed FSC, for all $z\in \observations$, we would have $\fsctransitionmap(n_i,z,z')=n_j$,  
	where $n_j$ is the node corresponding to the policy $\sigma_j$ added in the $j^{\textrm{th}}$ iteration.
	
	\emph{Case 3:} From observation $z$, the policy takes an action $a$ and then depending on the subsequent observation $z'$, it can switch to another policy $\sigma_j$. 
	In that case, in the constructed FSC, for all $z'\in \observations$, we would have $\fsctransitionmap(n_i,z,z')=n_j$,
	where $n_j$ is the node corresponding to the policy $\sigma_j$ added in the $j^{\textrm{th}}$ iteration. We also have $\fsclabelmap(n_i,z)=a$.

The iterative procedure concludes once a winning policy is identified from the initial observations. 
The controller created by this approach would still have some nodes non-reachable from the initial node.
We remove these irrelevant nodes by graph post-processing.

For more details of the algorithm, see \cite{JungesJS20}.
Here we will give a bit more intuition about case 2.
Suppose at iteration $i$, we find a policy that is winning from any state with observation $z$.
Then at any later iteration, $j>i$,
upon observing $z$, the agent can just switch to the previously discovered policy that is winning from $z$. This gives us the following fact:
\begin{restatable}{observation}{chainFSC}\label{obs:chain-FSC}
For any winning observation $z\in\observations$, there exists an iteration $i_z$ where we find a policy that is winning from any state with observation $z$.
Then in the FSC generated by the iterative approach, for any $j>i_z$, for any $z'\in \observations$, $\fsctransitionmap(n_j,z',z)=n_{i_z}$. 
\end{restatable}
\paragraph{Skip Transitions and \skippp-FSC}
We introduce an alternative representation for the FSCs produced by the iterative approach, referred to as \skippp-FSCs where we exploit the structural characteristics as described in \cref{obs:chain-FSC}. We define a \skippp-FSC as follows:
\begin{definition}[\skippp-FSC]
	A \skippp-FSC is a tuple $\fsc_{\skippp}=(\fscstates,\fscinitstate, \fsctransitionmap_{\skippp}, \fsclabelmap_{\skippp})$ where
	$\fscstates$ is a finite set of nodes,
	$\fscinitstate\in \fscstates$ is the initial node,
	$\fsctransitionmap_{\skippp}\colon \fscstates\times \observations\times\observations\to \fscstates$ is a special transition function,
	and $\fsclabelmap_{\skippp}\colon \fscstates\times \observations\to \mdpactions\cup\{\skippp\}$ is a special action labeling.
\end{definition} 
\begin{algorithm}[t]
	\caption{Semantics of \skippp-FSC: single-step execution}
	\label{alg:skip_fsc_execution}
		\KwIn {\skippp-FSC $\fsc_{\skippp}=(\fscstates,\fscinitstate, \fsctransitionmap_{\skippp}, \fsclabelmap_{\skippp})$, POMDP $\pomdp =(\mdp, \observations, \observationmap)$, current node $n$ and observation $z$}
		
		\Repeat{$a \neq \skippp$}{
			$a \gets \fsclabelmap_{\skippp}(n,z)$\;
			$n \gets \fsctransitionmap_{\skippp}(n,z,z)$\; 
		}
		Execute $a$ and go to the next state with observation $z'$\;
		$n \gets \fsctransitionmap_{\skippp}(n,z,z')$ \;
\end{algorithm}

In state $s$, an agent receives observation $z = \observationmap(s)$. 
At current node $n$, if $\fsclabelmap_{\skippp}(n, z)=\skippp$,
the agent transitions to the node $n' = \fsctransitionmap_{\skippp}(n,z,z)$ and continues this process until reaching a node $n''$ with 
$\fsclabelmap_{\skippp}(n'', z)\neq\skippp$.
The agent takes the action $a=\fsclabelmap_{\skippp}(n'', z)$.
The state of the MDP is then changed to $s'$ according to the probability distribution. 
Then, based on the current and the next observation $z$ and $z'=\observationmap(s')$, the FSC updates to the new node $\fsctransitionmap_{\skippp}(n'',z,z')$. 
The semantics of this FSC is described in \cref{alg:skip_fsc_execution}.

\subsubsection{Translation to a \skippp-FSC}  Given an FSC $\fsc=(\fscstates,\fscinitstate, \fsctransitionmap, \fsclabelmap)$ satisfying \cref{obs:chain-FSC}, we translate to a \skippp-FSC $\fsc_{\skippp}=(\fscstates,\fscinitstate, \fsctransitionmap_{\skippp}, \fsclabelmap_{\skippp})$ as described in \cref{alg:skip_fsc_creation}.
\begin{algorithm}[h]
	\caption{Attractor-based FSC Policy to \skippp-FSC}
	\label{alg:skip_fsc_creation}
	\KwIn {Attractor-based FSC $\fsc=(\fscstates,\fscinitstate, \fsctransitionmap, \fsclabelmap)$}
	\KwOut{\skippp-FSC $\fsc_{\skippp}=(\fscstates,\fscinitstate, \fsctransitionmap_{\skippp}, \fsclabelmap_{\skippp})$}
	$\fsctransitionmap_{\skippp} \gets \fsctransitionmap$; $\fsclabelmap_{\skippp} \gets \fsclabelmap$\;
	\For{winning observation $z'$}{
		\For{$j>i_{z'}$ and $z\in\observations$} { 
			Remove $(n_j,z,z')\mapsto n_{i_{z'}}$ from $\fsctransitionmap_{\skippp}$\;
			\For{$j>k>i_{z'}$}{
				$\fsctransitionmap_{\skippp}(n_k,z,z')\gets n_{k-1}$\;
				$\fsclabelmap_{\skippp}(n_k,z)\gets \skippp$\;
			}
		}
	}
	\Return $\fsc_{\skippp}=(\fscstates,\fscinitstate, \fsctransitionmap_{\skippp}, \fsclabelmap_{\skippp})$
\end{algorithm}
The algorithm iterates over all winning observations $z'$, modifying the transitions and action labels to incorporate the skip mechanism. 
Specifically, for each node 
$n_j$ with index greater that $i_{z'}$ (defined in \cref{obs:chain-FSC}), the direct switching to $n_{i_{z'}}$ is removed and replaced with $(j-i_{z'})$ single-step \skippp-transitions.

\begin{example}
	Consider an example where the attractor-based policy synthesis constructs an FSC with ten nodes $n_0, n_1, \dots, n_9$, where the node $n_i$ was added at the $i^{th}$ iteration. $n_0$ corresponds to the earliest discovered stationary policy. For each iteration $j > 0$, the newly synthesized subpolicy may need to switch to one of the earlier nodes $\{n_0, \dots, n_{j-1}\}$ depending on the posterior observation $z'$. We compare the resulting standard FSC with its \skippp-FSC variant.
	
	For each node $n_j$, the transition function $\delta(n_j, z, z')$
	must distinguish between up to $j$ different target nodes $\{n_0, \dots, n_{j-1}\}$, one for each possible winning observation $z'$. Hence, across all ten nodes, the total number of distinct next-node (in the worst case) labels to encode is $\Sigma_{j=1}^{9} j = 45.$ This yields large decision trees with many distinct leaf labels.
	
	In our $\skippp$-FSC, we replace each direct jump by a one-step connection
    to the immediate predecessor:
    $\gamma_{\text{skip}}(n_j, z) = \textsf{skip}$ and $\delta_{\text{skip}}(n_j, z, z') = n_{j-1},
    $
    for all winning posterior observations $z'$ and all observation $z$.
    When the \textsf{skip} action is played, the controller
    moves to $n_{j-1}$ and repeats this process until it reaches a node
    with a non-skip action.
    Consequently, each $n_j$ has two possible
    successor ($n_j$ and $n_{j-1}$) for all winning observations,
    reducing the number of distinct transition labels to $2\times 9 = 18$, a $45 \!\to\! 18$ reduction (about $2.5\times$ fewer labels).
    Thus, the DTs representing transitions become substantially smaller.
\end{example}

We prove that the \skippp-FSC generated in this way would represent the same policy as the original FSC:
\begin{restatable}{theorem}{skipFSC}\label{thm:skip-fsc}
	An FSC $\fsc$ generated by the iterative approach, and the \skippp-FSC $\fsc_{\skippp}$ created in \cref{alg:skip_fsc_creation} from it represents the same policy.
\end{restatable}

\begin{figure*}[t]
	\centering
	\begin{subfigure}[]{0.49\textwidth}
		\centering
		\includegraphics[width=\textwidth]{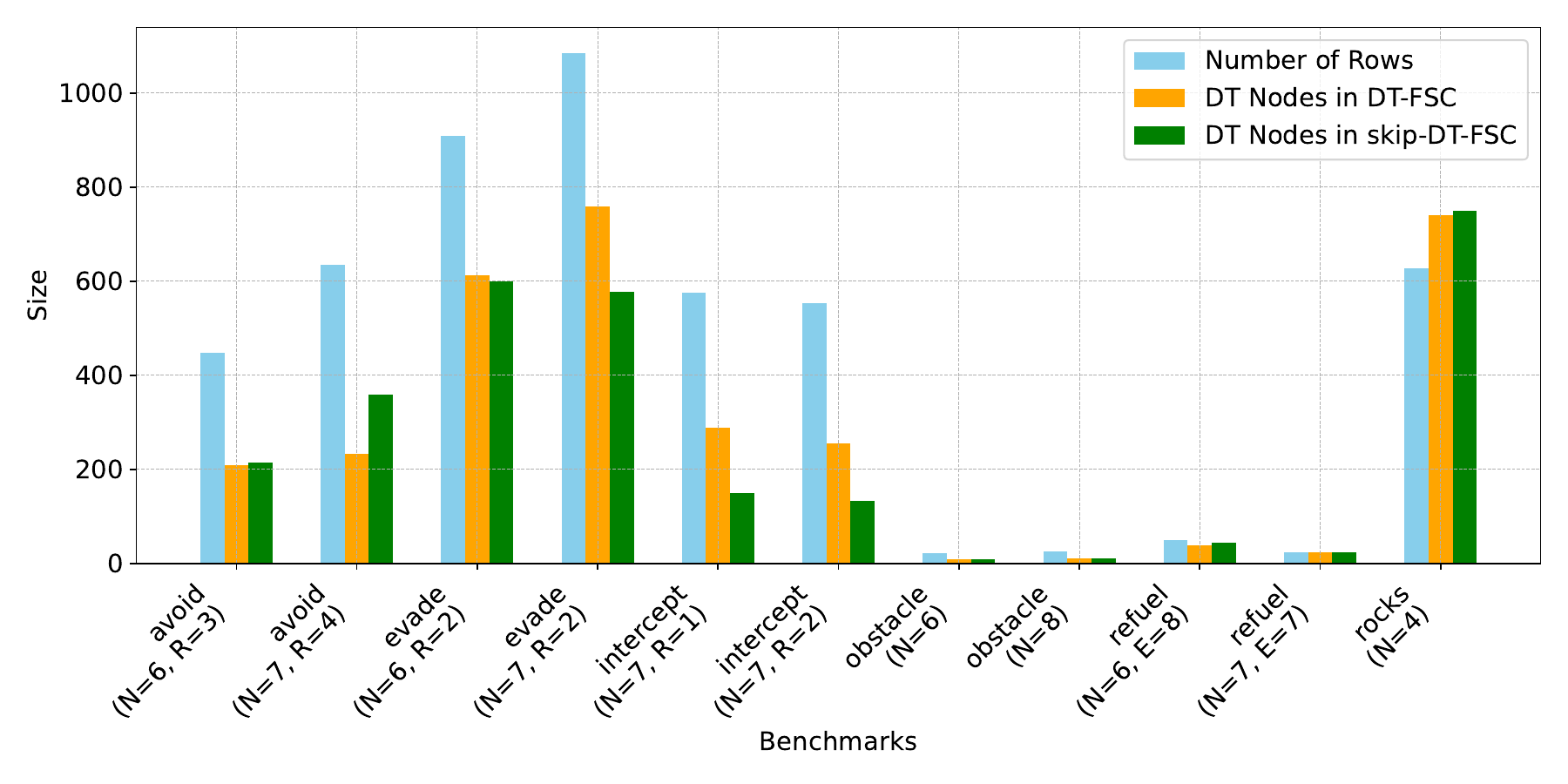}
		\caption{Different representation sizes of the action mapping $\gamma$. \emph{DT-FSC and \skippp-DT-FSC representations are respectively 1.7 and 1.87 times smaller than the tables.}}
		\label{fig:action_mapping}
	\end{subfigure}
	\hfill
	\begin{subfigure}[]{0.49\textwidth}
		\centering
		\includegraphics[width=\textwidth]{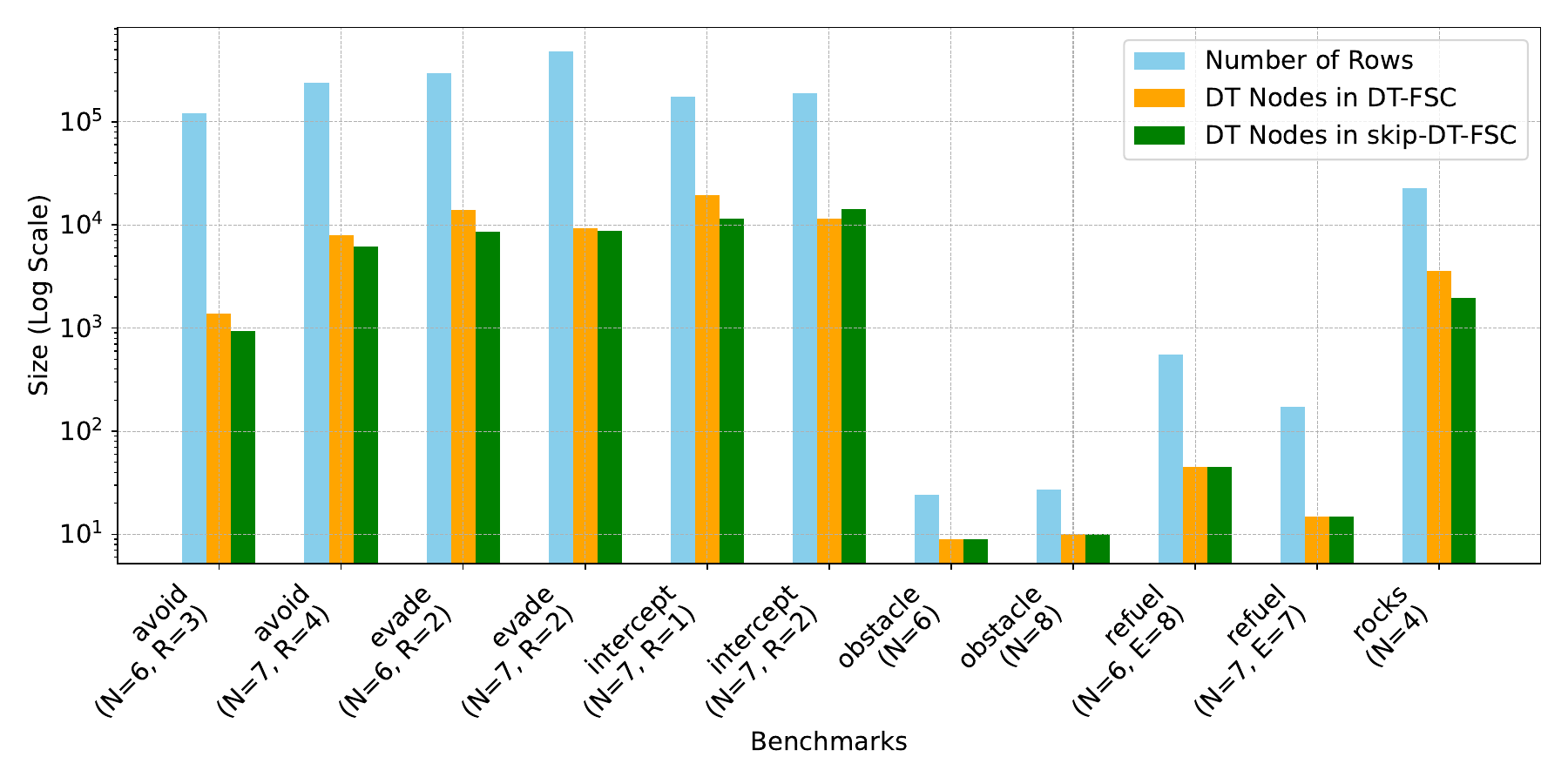}
		\caption{Different representation sizes of the transition function $\delta$. \emph{DT-FSC and \skippp-DT-FSC representations are respectively 13.5 and 16.37 times smaller than the tables.}}
		\label{fig:transition_functions}
	\end{subfigure}
	\vspace*{-0.4cm}
	\caption{{Comparison of the sizes of (posterior-aware) FSCs with our (posterior-aware) DT-FSCs for benchmarks obtained from~\protect\cite{JungesJS20}. \textbf{Log scale is used in \cref{fig:transition_functions}} for easier visualization of values spanning different magnitudes.}} 
	\label{fig:results}
	\vspace*{-0.25cm}
\end{figure*}

\paragraph{Translation from \skippp-FSC to \skippp-DT-FSC}
Finally, we apply Algorithm~\ref{alg:dt_fsc_creation} to the \skippp-FSC and represent its components using DTs, resulting in a \skippp-DT-FSC---essentially a DT-FSC with skip, represented using $\{\DT_{\fsctransitionmap_{\skippp},n}\}_{n\in \fscstates}$ and $\{\DT_{\fsclabelmap_{\skippp},n}\}_{n\in \fscstates}$.

\begin{remark}
	\emph{For these FSCs, the DT-FSC matches the original policy on all reachable observations. Differences may occur only on unreachable inputs, which do not affect the performance.}
\end{remark}

\section{Experimental Evaluation}
\label{sec:eval}

We adopt the viewpoint that explainability can be proxied by the size of the DTs. Compactness is straightforward to measure using this proxy; however, we further support explainability through a few examples and detailed case studies, demonstrating the interpretability and intelligibility provided by our representation. In particular, we answer the following questions: \emph{Q1): Does our approach make the policies more compact in terms of the size?} \emph{Q2) Does introducing $\skippp$-transitions improves the compactness of policies?}

\paragraph{Compact Representation of Policies.}

When converting an FSC to a DT-FSC, the number of nodes remains the same.
What changes is how the transition function $\delta$ and 
the action mapping $\gamma$ are represented.
For DT-FSCs, instead of representing them in a tabular form, we represent them as a more concise DT.
To show the quantitative improvement,
we compare the number of rows in the table and the number of nodes in the created DT.

\paragraph{Implementation Details.} 
We consider two sets of benchmarks:
(1) For the almost-sure reachability objectives, we use the all the benchmarks from~\cite{JungesJS20};
(2) for the general objectives, we take the benchmarks from \cite{bork2024learning} which includes models with objectives optimizing quantitative reachability or expected total reward to the target.  
In the first case, we modified the model checker \storm~\cite{storm} to get the optimal policies as an FSC. 
For the general case, we use the implementation in \cite{bork2024learning}.
Our prototype is implemented in Python, which processes the FSCs and uses dtControl~\cite{dtcontrol2} with the default settings to create the DTs. \emph{A docker image is publicly available at~\url{https://doi.org/10.5281/zenodo.17304441}.}

\paragraph{Results.}
\Cref{fig:results} summarizes the results for the benchmark set (1).
Our approach achieves significant reduction in size 
both in the case of\textbf{ 
the transition function (by a factor of 13.5 on average) and 
the action mapping (by a factor of 1.7 times on average)}---answering Q1 (see \cref{tab:results} in~\cref{app:experiments} for details).
For the benchmark set (2), the representation size of 
\textbf{the transition function is also reduced by a factor of 5.54 on average and 
the action mapping is reduced by a factor of 1.66 times on average }(see \cref{tab:general-case} in~\cref{app:experiments}). 
The action mapping tables are partial and depend only on current observations, while transition functions depend on both current and next observations. This makes action mapping tables smaller in our experiments and explains why DTs save more space in transition functions than in action mappings.
For the benchmarks in set (1), \textbf{the \skipp 
transition reduces the representation size of the
transition function by a factor of 16.37 on average and action selection by 1.87}---answering Q2 
(see \cref{tab:results} in~\cref{app:experiments}).

\paragraph{Time.} Policy synthesis times vary significantly depending on the benchmark,
with larger examples taking over an hour, while DT-FSC translation consistently requires
only a few seconds. 
For instance, across all benchmarks in set(1), policy synthesis took a total of $94$ minutes,
whereas DT-FSC creation was completed in less than $6$ minutes, highlighting the efficiency of 
DT-FSCs as a lightweight postprocessing step (see \cref{tab:time} in~\cref{app:experiments}).

\section{Explainability using Case Studies}
While direct quantitative measurement of explainability is challenging,
we aim to assess the explainability of our DT-FSC policies through illustrative case studies. 
Specifically, we aim to address the question: 
\emph{Does our proposed approach yield more explainable and practically useful policies?} 
Below, we detail two case studies to demonstrate the explanatory capabilities of our modular DT-FSC representation. 
For an additional case study, namely \emph{`Obstacle'}~\cite{JungesJS20} and further details, see Appendix~\ref{app:case_studies}.

\begin{figure*}[t]
	\centering
	\begin{minipage}{0.29\textwidth}
		\centering
		\includegraphics[scale=1.07]{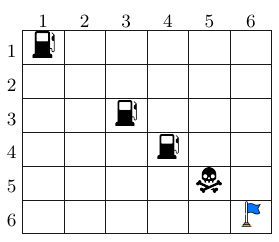}
	\end{minipage}\hfill
	\begin{minipage}{0.38\textwidth}
		\centering
			\includegraphics[width=0.8\textwidth]{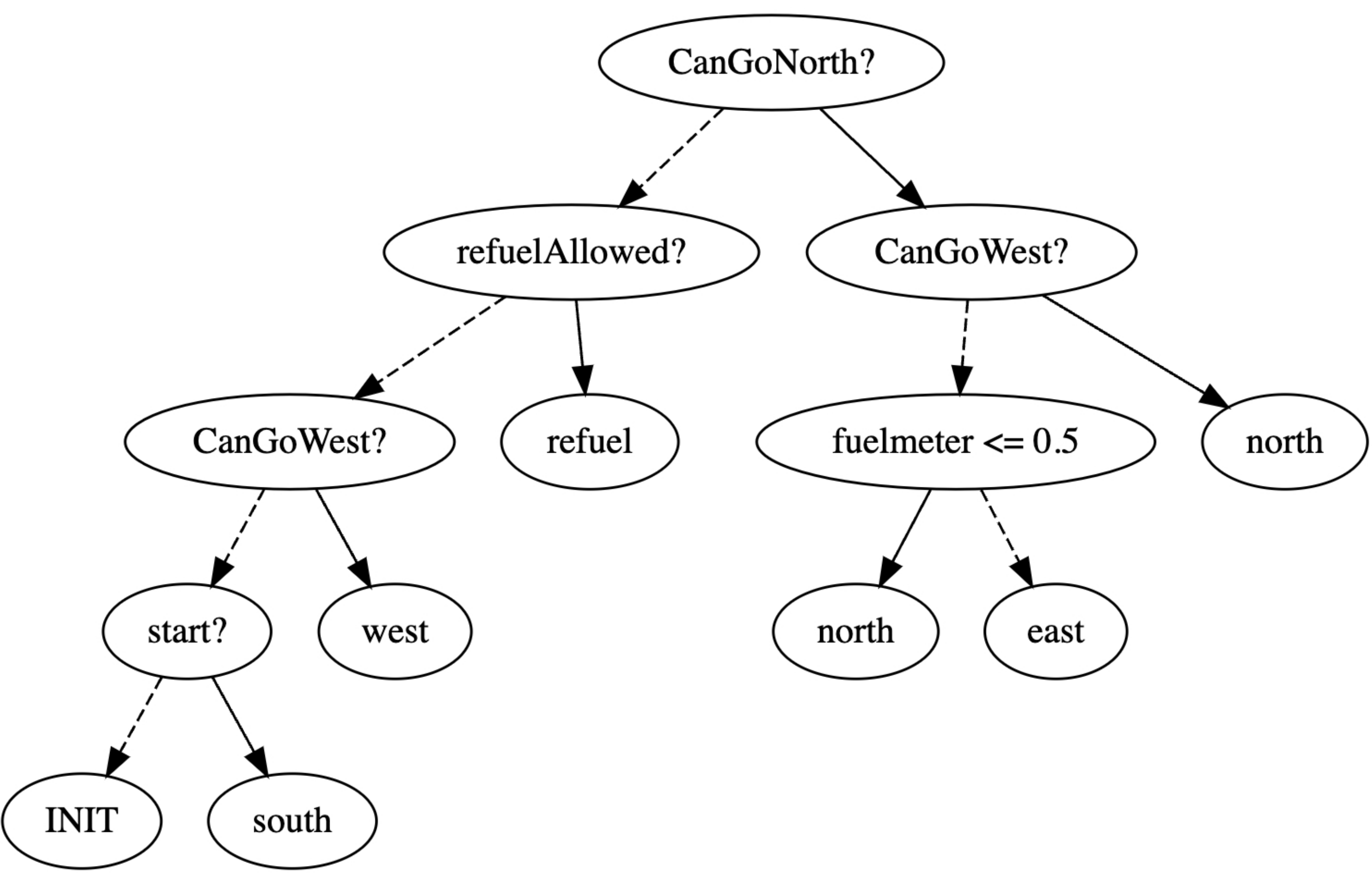}
	\end{minipage}\hfill
	\begin{minipage}{0.27\textwidth}
		\centering
			\includegraphics[scale=0.1]{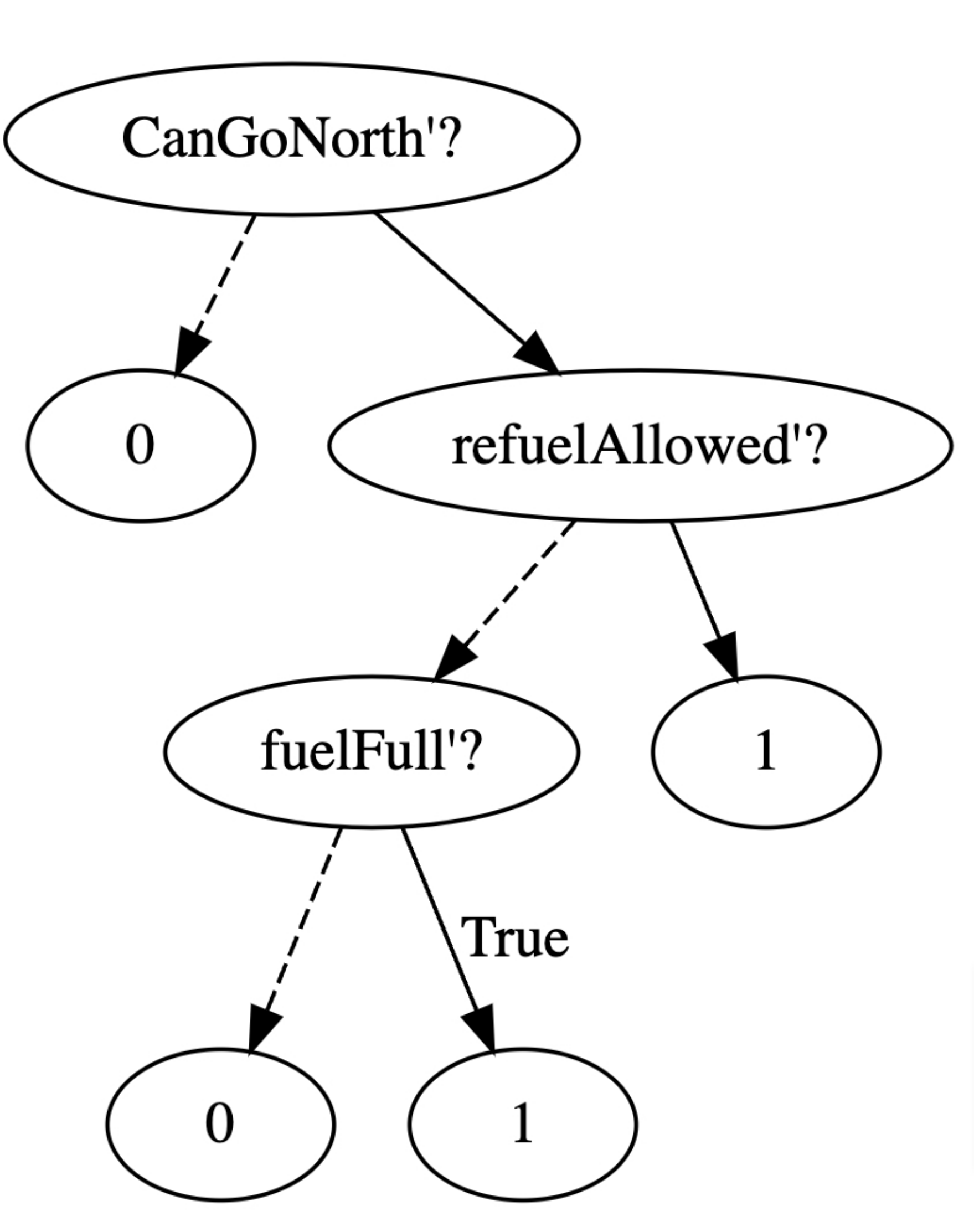}
	\end{minipage}
	\caption{The refuel model for $6 \times 6$ grid (Left), the DT representing the policy (Center), and the DT representing the transitions (Right) together illustrate the behavior of the initial node of the DT-FSC. Here, the solid edges in the DT are used to denote \texttt{true}, dotted to denote \texttt{false}. Initially, the rover cannot move north or west, and by following the decision tree, it first moves south. After this, moving north becomes possible, and the rover then follows the conditions in the DT to move east. If it has not yet reached the refueling station at $(3, 3)$ i.e., refueling is still not possible -- it continues moving north until it encounters a wall.\emph{Thus, the initial policy suggests cyclically repeating the sequence of actions (south, east, (north)+, (west)+) where (north)+ and (west)+ corresponds to a sequence of that action until a wall is reached. This allows the agent to get back to its initial state in the corner to refuel or eventually reach $(3,3)$ safely. It changes to a different node in the FSC once it refuels in $(3,3)$.}}
	\label{fig:combined-refuel}
\end{figure*}

\subsection{Refuel} 

We consider a gridworld-based model that was introduced in \cite{dietterich1998maxq} and modeled later in \cite{JungesJS20}.
A rover must travel from one corner to the diagonally opposite corner in an \(6 \times 6\) grid while avoiding an obstacle.
The rover can observe (i)~if the current location is \texttt{start} (i.e., at (1,1)), \texttt{bad} (i.e., at (5,5)), or \texttt{good} (i.e., at (6,6)), (ii)~\texttt{fuelmeter}: a meter that provides an abstract idea of the current fuel level, (iii)~\texttt{refuelStation?}: if it is at a refueling station or not (i.e., it is in at a location where it can refuel), (iv)~\texttt{fuelFull?}: if can be refueled or not, and
(v) \texttt{CanGo\{Direction\}} i.e., the cardinal directions in which it can move based on grid-walls.
The rover can choose to move one step in any of the directions (unless there is a wall).
The grid is slippery, so there is a possibility to slip making it take 2 steps.
The rover starts with $8$ units of fuel. It consumes one unit of fuel for each step (irrespective of whether it slipped or not) and can fully refuel at the refueling stations located at specific coordinates within the grid.
The objective for the rover is to reach a \texttt{good} location while avoiding \texttt{bad} states and preventing fuel depletion.

\paragraph{Explanation of DT-FSC Policies for Refuel.}
 
Unlike the tabular representation of FSCs, where the mapping between observations, actions, and transitions appears as large, unstructured lookup tables (with $50$ rows describing which action to take and $555$ rows describing whether to switch the policy or not), the DT-FSC organizes the same information hierarchically using interpretable predicates. The DTs are significantly smaller (for example, more than $12$ times smaller in the case of transition representation). Looking at the DTs, we can thus understand the policy by tracing how each observation leads to an action or memory transition by traversing the DT, revealing the controller's decision logic step by step.
We explain the policy in~\cref{fig:combined-refuel} (Center) and transitions in~\cref{fig:combined-refuel} (Right) for the initial node of DT-FSC and sketch the remaining policy.
The policy suggests cyclically repeating the sequence of actions (south, east, (north)+, (west)+) in order where (north)+ and (west)+ corresponds to a sequence of that action until a wall is reached.
with the rover refueling by playing \texttt{refuel} whenever possible.
This way, the rover will eventually reach the second refueling station at $(3,3)$, a subgoal in the iterative procedure. Once refueled at the second refuel station, the memory location switches to the next one (see the DT in~\cref{fig:combined-refuel} (Right)), directing the rover to move south. After reaching the southernmost cell, the policy then suggests the rover continue east, ultimately reaching the target while avoiding the bad state.

\subsection{Planning Treatment of Heart Diseases} 
We model the diagnosis and treatment of ischemic heart disease as described in \cite{hauskrecht2000planning}.
The doctor diagnoses and treats patients potentially suffering from coronary artery disease (CAD) or myocardial ischemia (MI). 
The model accounts for cases of harmless chest pain, which do not require treatment and the risks of complications arising from diagnostic tests and treatment procedures.
The diseases (CAD and MI) are represented by hidden state variables,
while the symptoms (e.g., chest pain) are observable. 
The doctor can use several diagnostic tools with different detection rates: an electrocardiogram (EKG) test, stress test and angiogram.  
The test results are observable and reveal more information about the hidden state variables 
and can choose between non-invasive medication, 	
or invasive procedures like coronary artery bypass grafting (CABG) and percutaneous transluminal coronary angioplasty (PCTA). 
CABG is generally more effective but riskier, while PCTA is less invasive, but its success depends on the presence of complications. 

Waiting may resolve harmless symptoms but risks worsening serious conditions. 
Conducting tests, treatments, and waiting incur some costs.
These costs are higher if treatments are administered without proper diagnosis.
Additional costs are incurred if complications arise.
We modeled this POMDP in the \prism~language~\cite{prism} and used the FSC learning technique  with the default configuration implemented in \cite{bork2024learning} to create a $4$ node FSC.

\paragraph{Explanation of DT-FSC Policies for heart diseases.}

We provide only a partial overview of the policy here. 
Initially, as described in \cref{fig:heart-policy-0} in Appendix~\ref{app:case_studies},
the policy suggests \texttt{EKGTest}, and depending on the results, advises either additional tests or medication. It suggests going for an invasive procedure only when the stress test is positive. 
In the later iterations (refer to~\cref{app:case_studies}),
it suggests randomized policies in the cutoff points. 
While these policies can be challenging to interpret, they offer some insight into possible actions a doctor might take. 
For instance, if the policy assigns a higher probability to choosing \texttt{pcta} over \texttt{medicine}, then PCTA would be preferred in that scenario.
Therefore, for complex models like this, a hybrid approach combining DT-FSC with human expertise can offer the needed explainability and an effective policy.

\section{Conclusion}
Our method profits from the graphical representation of FSCs and combines it with the DT representation of (i) the action selection in the memoryless parts of the policy and (ii) the labeling of the transitions in the FSC.
Together this provides a modular and flexible framework, able to accommodate, \textit{e.g.}, richer transition functions (taking more observations as arguments) or richer actions (of randomized policies).
Not only does it allow for better explainability, but also for better diagnostics of where the complexities arise, \textit{e.g.}, in which situations the posterior observation is really needed or under which constraints randomization occurs (and thus a validation is possibly in place).
This could become even more efficient when the modeling process provides the observation and state information in a more structured way, \textit{e.g.}, as values of several sensors rather than an identification number.

\begin{acks}
    This work has been supported by the DFG project {GOPro} (427755713), MUNI Award in Science Humanities grant (MUNI/I/1757/2021), the ERC project {InOVationCS} (101171844) and the RSS Scheme (NRF-RSS2022-009) by NRF, Singapore.
\end{acks}

\balance
\bibliographystyle{ACM-Reference-Format} 
\bibliography{ref}

\newpage
\appendix

\section{Appendix}
\label{app:additional_details}

\subsection{Additional Details of Section~\ref{sec:solution}}
\label{app:illustrative_example}
\begin{figure}[!h]
	\centering
	\includegraphics[width=\columnwidth]{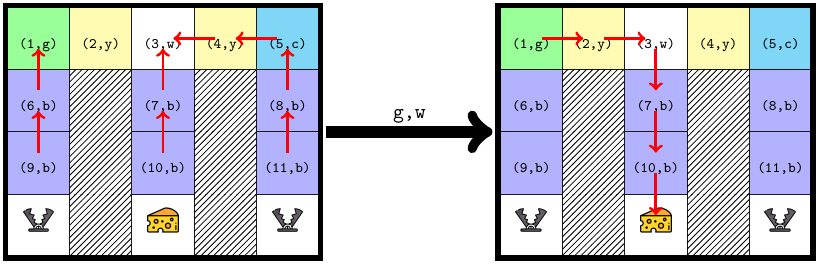}
	\caption{Policy of $n_0$ (Left) and $n_1$ (Right) in Maze and the transitions on the \textbf{bold} arrow. The numbers are to indicate different states and the letters are used for observations. Same letter for cells implies that the mouse can not distinguish between those states.}
	\label{fig:maze-policy}
\end{figure}

\cref{fig:maze-policy} shows the policy for~\cref{ex:maze}. For observations $\mathsf{b, y}$, policies are defined in both memory states, while $\mathsf{g \text{~and~} w}$ trigger a switch.

\subsection{Proof of Correctness for FSC with Skip}
\label{sec:skip-fsc-proof}
For our new \skippp-FSC, we first recall the following observation.
\chainFSC*
Next we formally prove the correctness of \Cref{alg:skip_fsc_creation}.
\skipFSC*
\begin{proof}
	We show that for any observation sequence $\rho$ in $\observations^*$, 
	starting from the initial node $\fscinitstate$, both the original $\fsc$ and the $\fsc_{\skippp}$ will reach the same node and suggest same action. This is proven by induction on $|\rho|$.
	
	\noindent\textbf{Base case :} $\rho=\observationmap(\mdpinitstates)$. We stop the iteration when $\observationmap(\mdpinitstates)$ is in the winning region. So both the FSCs would suggest $\fsclabelmap(\fscinitstate,\observationmap(\mdpinitstates))$.
	
	\noindent\textbf{Induction step :}
	Assume the claim  is true for some observation sequence $\rho$,
	meaning we reach the same node $n$ in both $\fsc$ and $\fsc_{\skippp}$.
	Now consider an extended observation sequence $\rho\cdot z$.
	Let $\fsctransitionmap(n,\last(\rho),z)=n'$. There are two cases to consider.
	\begin{itemize}
		\item [Case 1:] $n'=n_{i_z}$ where $i_z$ is the index of the node defined in \cref{obs:chain-FSC}.  
		From \cref{alg:skip_fsc_creation}, we have $\fsclabelmap_{\skippp}(n,z)=\skippp$ for all nodes from $n$ to $n_{i_z}$. 
		Then, for the observation sequence $\rho\cdot z$, we reach the same node $n_{i_z}$ in both $\fsc$ and $\fsc_{\skippp}$. 
		Also, as we are not adding any $\skipp$ actions from $z$ at $n_{i_z}$, we have $\fsclabelmap_{\skippp}(n_{i_z},z)=\fsclabelmap(n_{i_z},z)$.
		\item [Case 2:] $n'\neq n_{i_z}$
		This implies that  $\fsctransitionmap(n,\last(\rho),z)=\fsctransitionmap_{\skippp}(n,\last(\rho),z)=n'$
		and $\fsclabelmap_{\skippp}(n_{i_z},z)\neq \skippp$. So $\fsclabelmap_{\skippp}(n_{i_z},z)=\fsclabelmap_{\skippp}(n_{i_z},z)$.
	\end{itemize}
	
Thus, the original FSC $\fsc$ and the skip-FSC $\fsc_{\skippp}$ both reach the same node and suggest the same action for any observation sequence $\rho$.
Therefore, $\fsc$ and $\fsc_{\skippp}$ represent the same policy.
\end{proof}

\subsection{Additional details on Case Studies}
\label{app:case_studies}
\begin{figure*}
	\centering
	\includegraphics[width=0.8\textwidth]{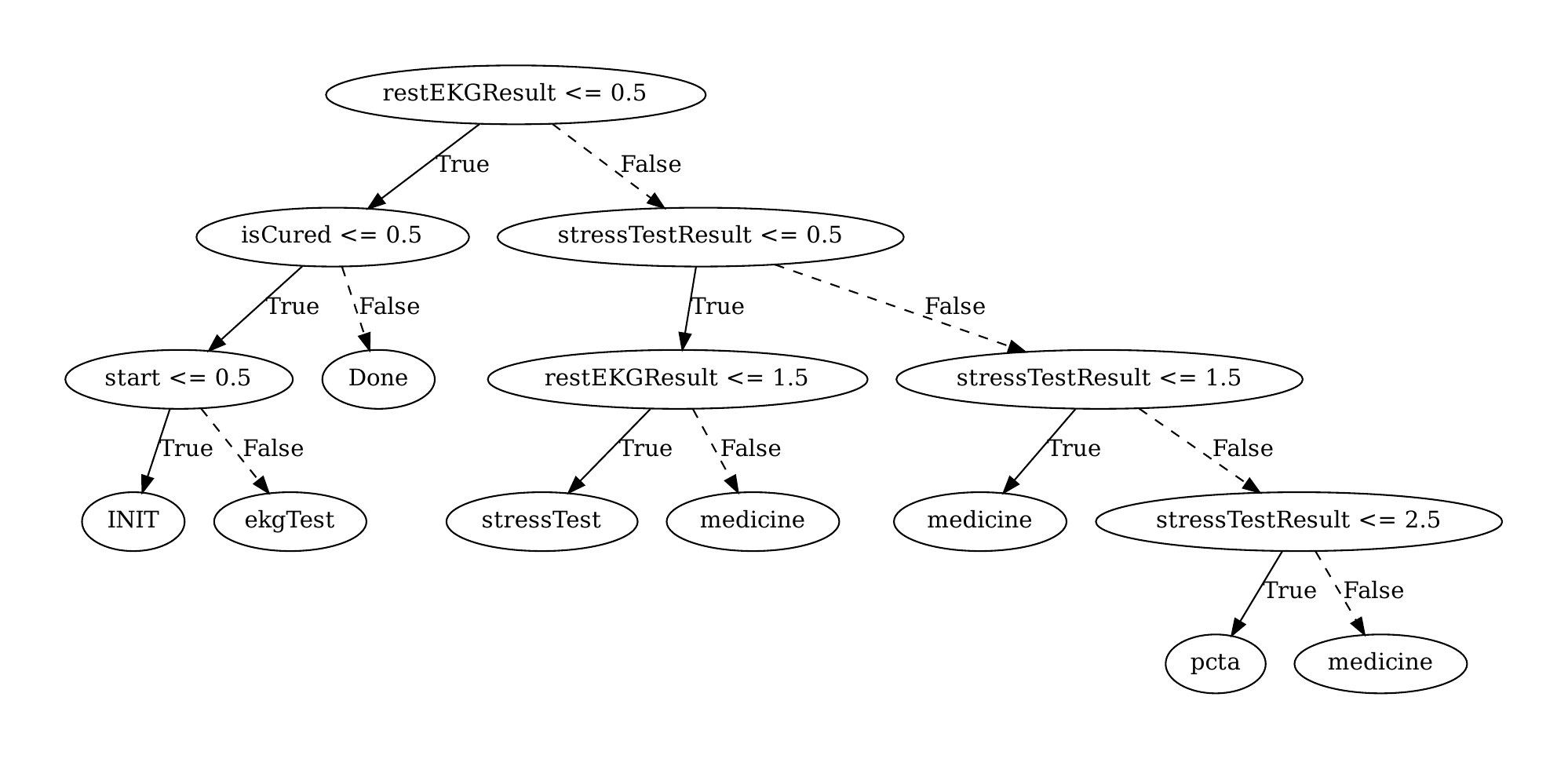}
	\caption{DT representation of the stationary policy at node $0$ for treatment of heart diseases}
	\label{fig:heart-policy-0}
\end{figure*}

\begin{figure*}
	\centering
	\includegraphics[width=0.8\textwidth]{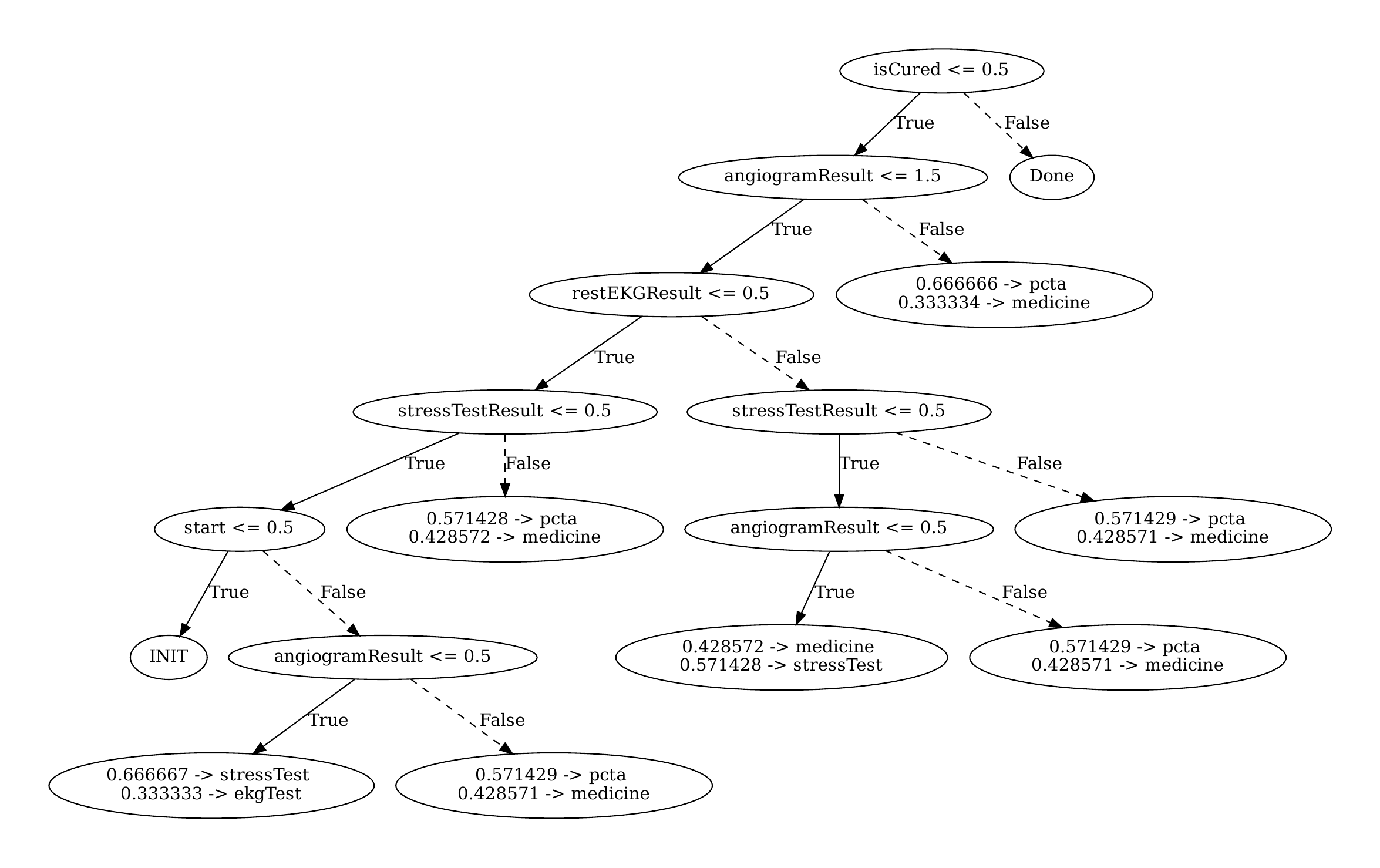}
	\caption{DT representation of the stationary policy at node $3$ for treatment of heart diseases}
	\label{fig:heart-policy-3}
\end{figure*}

\begin{figure*}
	\centering
	\includegraphics[width=0.8\textwidth]{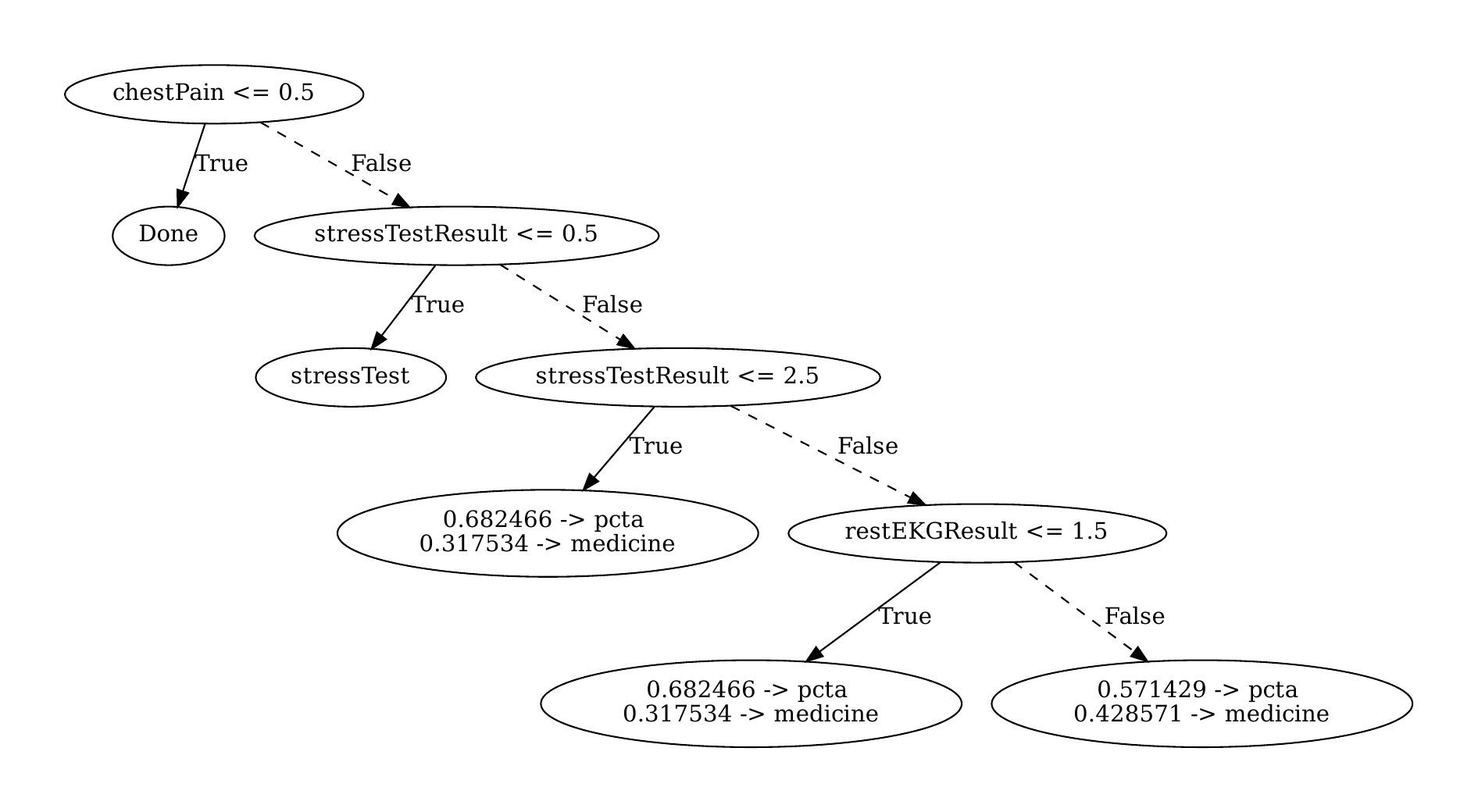}
	\caption{DT representation of the stationary policy at node $9$ for treatment of heart diseases}
	\label{fig:heart-policy-9}
\end{figure*}

\begin{figure*}
	\centering
	\includegraphics[width=0.8\textwidth]{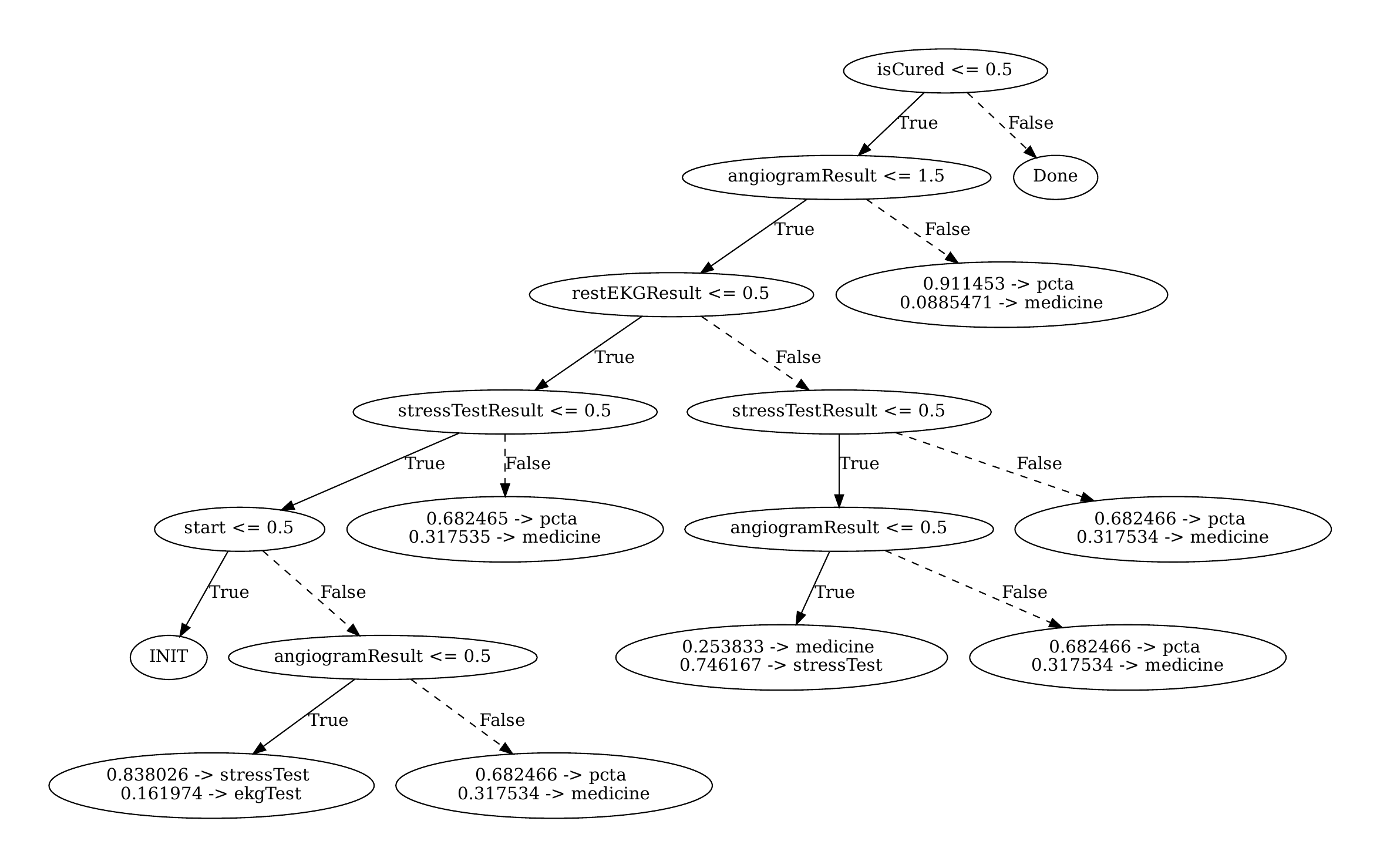}
	\caption{DT representation of the stationary policy at node $11$ for treatment of heart diseases}
	\label{fig:heart-policy-11}
\end{figure*}

\begin{figure*}[t]
	\centering
	\begin{subfigure}{0.45\textwidth}
		\centering
	\includegraphics[width=\textwidth]{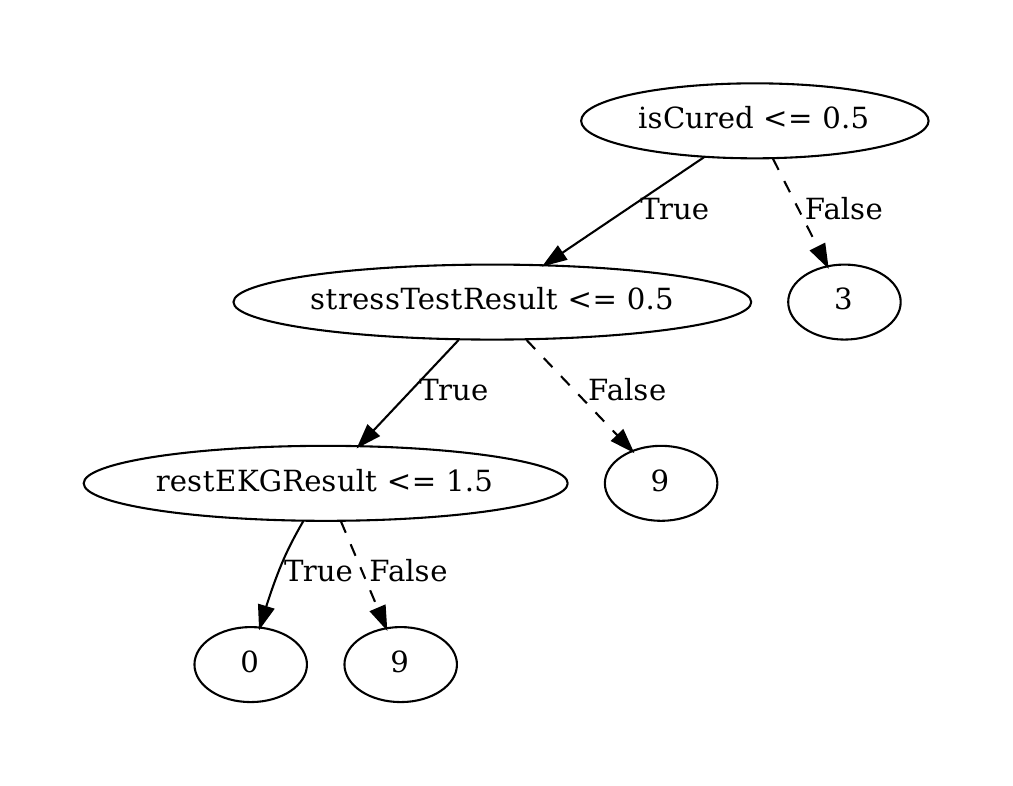}
	\caption{From node $0$}
	\end{subfigure}
	\begin{subfigure}{0.45\textwidth}
	\centering
	\includegraphics[width=0.5\textwidth]{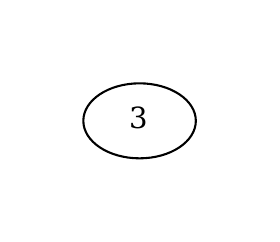}
	\caption{From node $3$}
	\end{subfigure}

	\begin{subfigure}{0.45\textwidth}
	\centering
	\includegraphics[width=\textwidth]{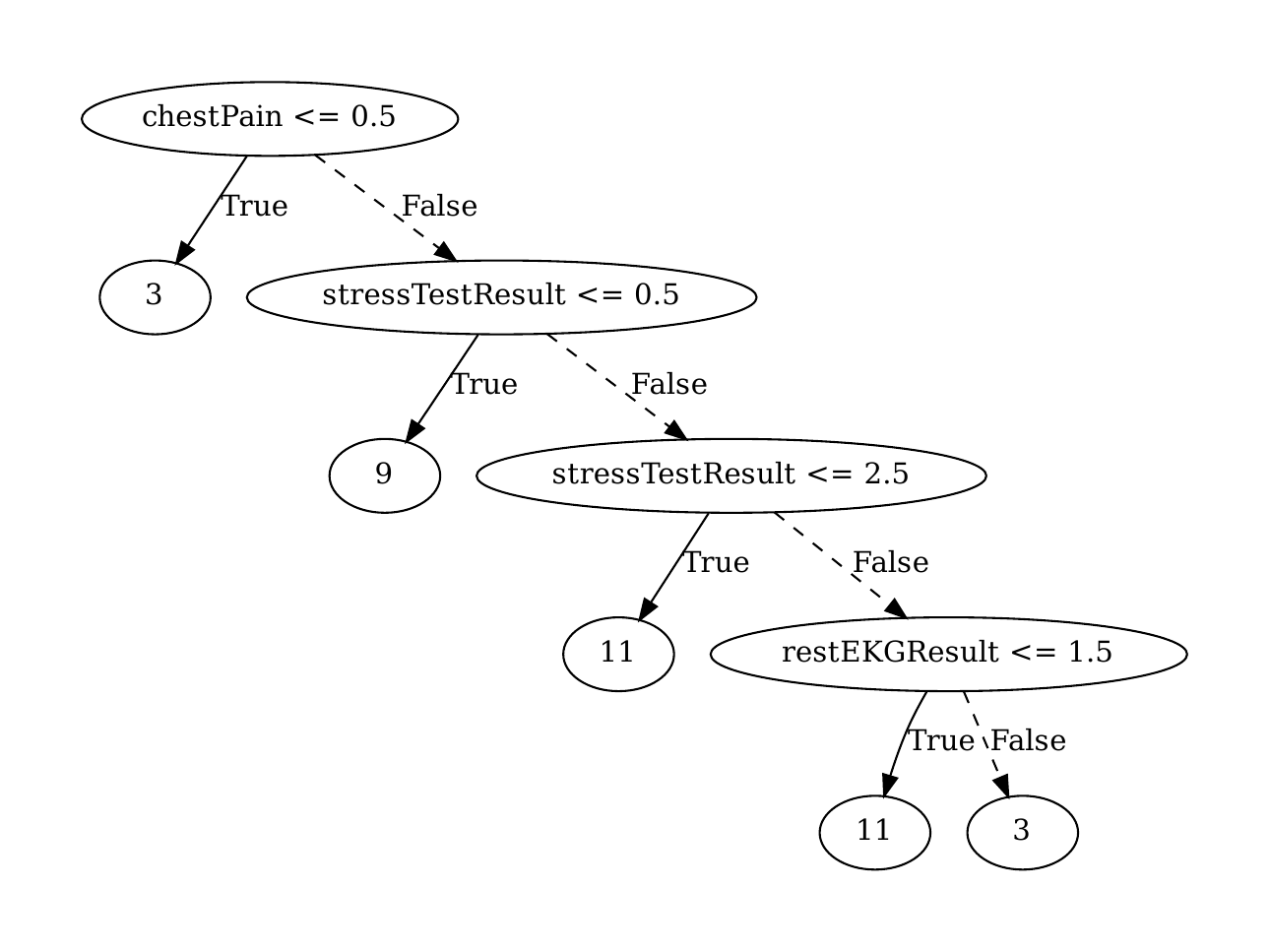}
	\caption{From node $9$}
	\end{subfigure}
	\begin{subfigure}{0.45\textwidth}
	\centering
	\includegraphics[width=0.5\textwidth]{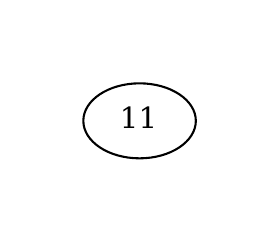}
	\caption{From node $11$}
\end{subfigure}

\caption{DT representation of the FSC transitions from different nodes in the FSC for treatment of heart diseases}
\label{fig:heart-transitions}
\vspace*{0.1cm}
\end{figure*}

\paragraph{Heart diseases.} We show all the policies (\cref{fig:heart-policy-0} to \cref{fig:heart-policy-11}) and transitions (\cref{fig:heart-transitions}) for the heart example. Observe that the next policies suggest randomizing between different procedures - which gives a hint based on probabilities what could be a better choice. However, it also shows that we need human intervention in order to take more rational decisions.

\begin{figure}[h]
	\centering
	\includegraphics[width=0.85\columnwidth]{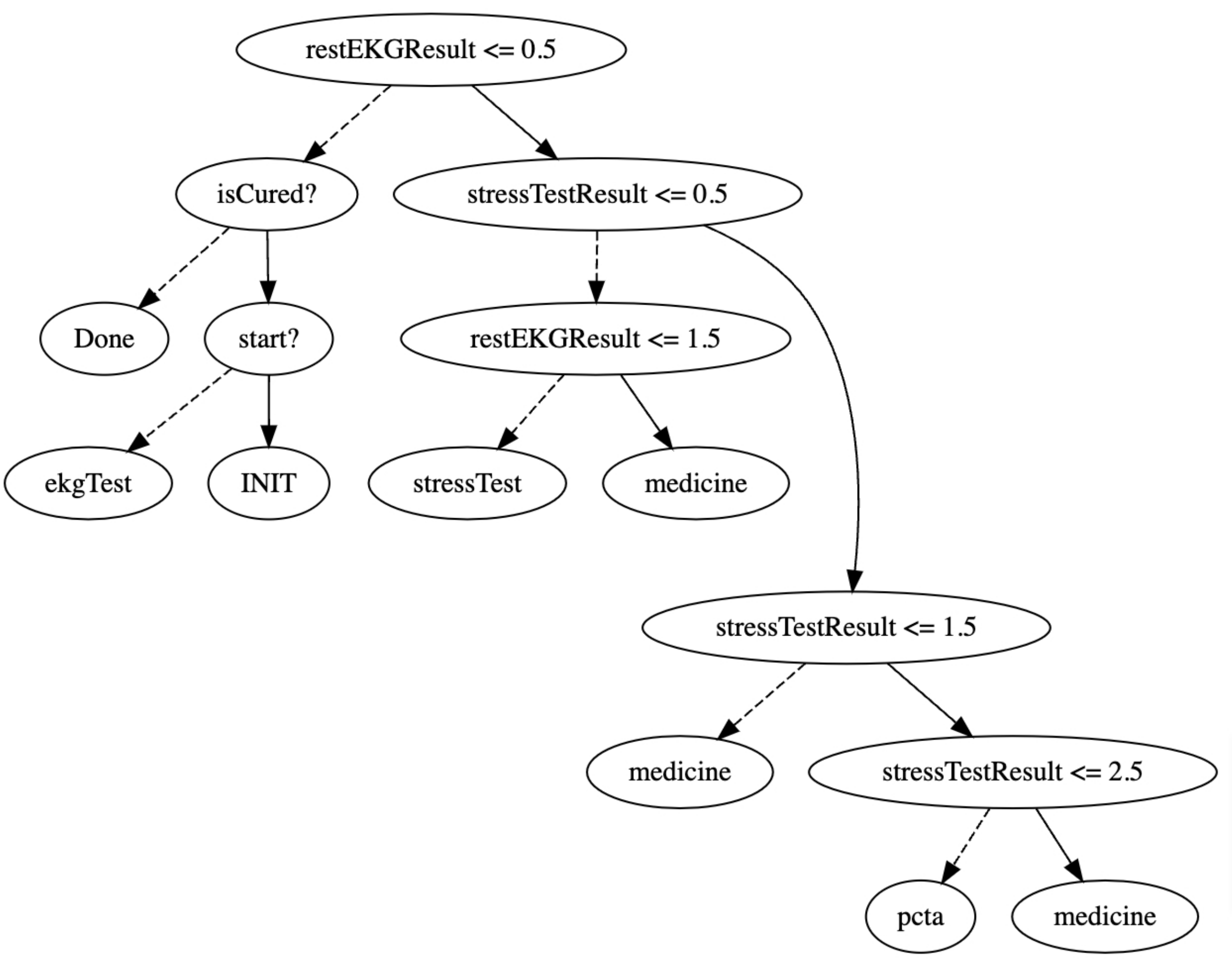}
	\caption{Decision tree representation of the initial heart disease treatment policy from the FSC. Test results: 0 = not done, 1 = negative, 2 = positive, 3 = inconclusive.}
	\label{fig:dts-action-selection-heart}
\end{figure}

\paragraph{Obstacle.} This case study involves a robot navigating through an environment with static obstacles to reach a designated exit. The environment is partially observable, adding complexity to the decision-making process. The robot begins in an initial state that is uncertain.

The robot has limited observational capabilities; it can only determine whether its current position is either an ``obstacle'' (an area that prevents further progress and penalizes the robot) or the ``exit'' (the target destination). This restricted sensory input mirrors realistic scenarios where robots operate with incomplete or noisy information about their surroundings.

\begin{figure}[H]
	\centering
	\includegraphics[width=0.8\columnwidth]{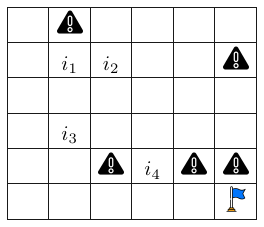}
	\caption{The obstacle case study with $6 \times 6$ grid, 5 obstacles, and 4 non-deterministic initial locations ($i_1,i_2,i_3,$ and $i_4$) of the robot.}
	\label{fig:obstacle}
\end{figure}
\paragraph{Explanation of DT-FSC Policies for Obstacle.} 
The DT-FSC policy generated by our approach demonstrates how the robot can navigate the grid safely, regardless of its starting position ($i_1, i_2, i_3$, or $i_4$). The initial stationary policy advises the robot to take a single step towards \texttt{south}. Afterward, the policy utilizes three memory nodes to guide the robot through a series of steps: first moving completely to the west playing \texttt{west}  action 3 times (note the initial position $i_4$), then taking three steps \texttt{south}, and finally continuing to the \texttt{east} until it reaches the target.
 Each memory location in the policy contains only a single decision tree (DT) node, which dictates the direction the robot should move. Essentially, the policy incorporates an implicit step-counting mechanism, as the robot can only observe whether it has reached the target or encountered an obstacle.

\subsection{Additional Details on Experimental Evaluation}
\label{app:experiments}
We present a summary of additional experimental results here. Additionally, we provide the data to generate these values. The instruction are in the \texttt{README.md} accompanied by this Appendix. Moreover, we provide a link to a publicly available docker image  which can be used to reproduce the results from scratch.

\paragraph{Effect of \skipp Actions.}
\cref{tab:results} demonstrates that when comparing the number of rows in a FSC in tabular form to a DT-FSC created from the \skippp-FSCs, the latter is more efficient. 
Specifically, the tables have $1.87$ times more rows as compared to the number of nodes in DT representation of the policies in the \skippp-DT-FSC and $16.37$ times more rows as compared to the number of nodes in DT representation of the transitions in the \skippp-DT-FSC.

\paragraph{Results for Benchmark Set (1).}

\begin{table*}[t!]
	\caption{Comparison of the sizes of (posterior-aware) FSCs with our (posterior-aware) DT-FSCs and \skipp-DT-FSC for benchmark set (1) obtained from~\protect\cite{JungesJS20}.}
	\label{tab:results}
	\centering
	\resizebox{\textwidth}{!}{
\begin{tabular}{l|r|rrrrr|rrrrr}
	\toprule
	\multirow{2}{*}{\textbf{Benchmark}} & 
	\multirow{2}{*}{\textbf{\#Nodes}} & 
	\multicolumn{5}{c|}{\textbf{Stationary Policy Size}} & 
	\multicolumn{5}{c}{\textbf{Transition Size}} \\ \cmidrule{3-12}
	& & \textbf{\#Rows} & \textbf{\#DT-} & \textbf{Ratio} & \textbf{\#Skip-} & \textbf{Ratio} & \textbf{\#Rows} & \textbf{\#DT-} & \textbf{Ratio} & \textbf{\#Skip-} & \textbf{Ratio} \\ 
	& & & \textbf{nodes} & \textbf{(FSC/} & \textbf{DT-} & \textbf{(FSC/} & & \textbf{nodes} & \textbf{(FSC/} & \textbf{DT-} & \textbf{(FSC/} \\ 
	& & & & \textbf{DT-FSC)} & \textbf{nodes} & \textbf{Skip-DT)} & & & \textbf{DT-FSC)} & \textbf{nodes} & \textbf{Skip-DT)} \\ \midrule
	avoid (N=6, R=3) & 9 & 447 & 209 & 2.14 & 215 & 2.08 & 121837 & 1381 & 88.21 & 945 & 129.01 \\ 
	avoid (N=7, R=4) & 3 & 634 & 233 & 2.72 & 359 & 1.77 & 240418 & 7927 & 30.32 & 6135 & 39.19 \\ 
	evade (N=6, R=2) & 31 & 908 & 613 & 1.48 & 599 & 1.52 & 296100 & 14003 & 21.14 & 8607 & 34.39 \\ 
	evade (N=7, R=2) & 30 & 1085 & 758 & 1.43 & 578 & 1.88 & 477329 & 9226 & 51.73 & 8818 & 54.15 \\ 
	intercept (N=7, R=1) & 5 & 575 & 289 & 1.99 & 149 & 3.86 & 176714 & 19551 & 9.04 & 11549 & 15.30 \\ 
	intercept (N=7, R=2) & 5 & 554 & 255 & 2.17 & 133 & 4.17 & 189017 & 11421 & 16.55 & 14157 & 13.35 \\ 
	obstacle (N=6) & 7 & 22 & 9 & 2.44 & 9 & 2.44 & 24 & 9 & 2.67 & 9 & 2.67 \\ 
	obstacle (N=8) & 8 & 25 & 10 & 2.50 & 10 & 2.50 & 27 & 10 & 2.70 & 10 & 2.70 \\ 
	refuel (N=6, E=8) & 5 & 50 & 39 & 1.28 & 45 & 1.11 & 555 & 45 & 12.33 & 45 & 12.33 \\ 
	refuel (N=7, E=7) & 3 & 24 & 23 & 1.04 & 23 & 1.04 & 172 & 15 & 11.47 & 15 & 11.47 \\ 
	rocks (N=4) & 52 & 627 & 740 & 0.85 & 750 & 0.84 & 22634 & 3578 & 6.33 & 1968 & 11.50 \\ \midrule
	\multicolumn{2}{l|}{\textbf{Geometric Mean}} & & & \textbf{1.7} & & \textbf{1.87} & & & \textbf{13.5} & & \textbf{16.37} \\ \bottomrule
\end{tabular}
}
\vspace*{0.3cm}
\end{table*}

For benchmark set (1), we compare the sizes of (posterior-aware) FSCs with our (posterior-aware) DT-FSCs for benchmarks obtained from~\protect\cite{JungesJS20}.
\cref{tab:results} describes the results.
\cref{tab:general-case} describes the results.
\paragraph{Results for Benchmark Set (2).}
For benchmark set (2), we generate DT-FSCs from the FSCs created in \cite{bork2024learning} for quantitative benchmarks, we achieve a significant reduction in transition sizes. 
\cref{tab:general-case} describes the results.

\begin{table*}[ht]
		\caption{Comparison of FSC sizes with our DT-FSC sizes for the benchmark set (2) obtained from~\protect\cite{bork2024learning}}
	\label{tab:general-case}
	\centering
	\resizebox{\textwidth}{!}{
		\begin{tabular}{l|r|rrr|rrr}
			\toprule
			\multirow{2}{*}{\textbf{Benchmark}} & \multirow{2}{*}{\textbf{\#FSC-nodes}} & \multicolumn{3}{c|}{\textbf{Stationary Policy size}} & \multicolumn{3}{c}{\textbf{Transition size}} \\ \cmidrule{3-8}
			&  & \textbf{\#Rows} & \textbf{\#Nodes} & \textbf{Ratio (Rows/Nodes)} & \textbf{\#Rows} & \textbf{\#Nodes} & \textbf{Ratio (Rows/Nodes)} \\ \midrule
			refuel-06 & 3 & 71 & 75 & 0.95 & 89 & 19 & 4.68 \\
			grid-large-30-5 & 1 & 39 & 39 & 1.00 & 40 & 1 & 40.00 \\ 
			problem-storm-extended & 61 & 246 & 65 & 3.78 & 247 & 63 & 3.92 \\ 
			grid-avoid-4-0 & 5 & 24 & 13 & 1.85 & 26 & 9 & 2.89 \\ 
			posterior-awareness & 4 & 20 & 8 & 2.50 & 20 & 12 & 1.67 \\ 
			grid-avoid-4-0 & 5 & 22 & 9 & 2.44 & 24 & 7 & 3.43 \\ 
			hallway & 18 & 198 & 206 & 0.96 & 198 & 144 & 1.38 \\ 
			4x3-95 & 9 & 71 & 63 & 1.13 & 71 & 55 & 1.29 \\ 
			stand-tiger-95 & 4 & 28 & 18 & 1.56 & 28 & 14 & 2.00 \\ 
			query-s3 & 16 & 104 & 72 & 1.44 & 104 & 66 & 1.58 \\ 
			grid-large-20-5 & 1 & 19 & 19 & 1.00 & 20 & 1 & 20.00 \\ 
			problem-storm & 3 & 14 & 7 & 2.00 & 15 & 5 & 3.00 \\ 
			rocks-16 & 2 & 2621 & 826 & 3.17 & 2774 & 4 & 693.50 \\ 
			query-s2 & 9 & 62 & 39 & 1.59 & 62 & 37 & 1.68 \\ 
			network-prio-2-8-20 & 1 & 4896 & 429 & 11.41 & 4912 & 1 & 4912.00 \\ 
			problem-storm-paynt-combined & 7 & 40 & 29 & 1.38 & 41 & 21 & 1.95 \\ 
			network & 5 & 29 & 23 & 1.26 & 29 & 11 & 2.64 \\ 
			milos-aaai97 & 3 & 24 & 13 & 1.85 & 24 & 5 & 4.80 \\ 
			refuel-08 & 5 & 113 & 115 & 0.98 & 136 & 43 & 3.16 \\ 
			hallway2 & 16 & 317 & 262 & 1.21 & 317 & 206 & 1.54 \\ 
			rocks-12 & 2 & 1541 & 542 & 2.84 & 1658 & 4 & 414.50 \\ 
			web-mall & 3 & 17 & 11 & 1.55 & 17 & 7 & 2.43 \\ 
			4x5x2-95 & 18 & 107 & 86 & 1.24 & 107 & 74 & 1.45 \\ 
			problem-paynt-storm-combined & 6 & 39 & 26 & 1.50 & 42 & 14 & 3.00 \\ 
			maze-alex & 16 & 119 & 114 & 1.04 & 122 & 84 & 1.45 \\ 
			grid-large-10-5 & 8 & 36 & 14 & 2.57 & 37 & 8 & 4.63 \\ 
			problem-paynt & 3 & 13 & 5 & 2.60 & 13 & 5 & 2.60 \\ 
			mini-hall2 & 3 & 25 & 25 & 1.00 & 25 & 15 & 1.67 \\ 
			refuel-20 & 4 & 194 & 196 & 0.99 & 227 & 22 & 10.32 \\ \midrule
			\multicolumn{2}{l|}{\textbf{Geometric Mean}} & & & \textbf{1.66} & & & \textbf{5.54} \\ \bottomrule
		\end{tabular}
	}
\end{table*}

\paragraph{Time.}

\begin{table*}[ht]
	\centering
	\caption{Comparison of time (in seconds) taken for synthesizing the policies and the time taken for creating the DT-FSCs as a postprocess step for the benchmarks obtained from~\protect\cite{JungesJS20}.}
	\label{tab:time}
	\resizebox{\textwidth}{!}{
		\begin{tabular}{l|rrr|rr}
			\toprule
			\textbf{Benchmarks}&\textbf{\#States}&\textbf{\#Transitions}&\textbf{\#Observations}&\textbf{Synthesis Time (sec.)} & \textbf{DT-FSC Creating Time (sec.)}\\
			\midrule
			avoid (N=6, R=3) &5976&14373&3300&57.65&20.56\\
			avoid (N=7, R=4) &13021&33949&8584&4715.34&23.31\\
			evade (N=6, R=2) &4232&28866&2202&144.97&70.18\\
			evade (N=7, R=2) &8108&57570&4172&541.93&77.04\\
			intercept (N=7, R=1) &4705&18049&2002&60.72&22.20\\
			intercept (N=7, R=2) &4705&18049&2598&55.24&20.99\\
			obstacle (N=6) &37&224&4&0.17&10.51\\
			obstacle (N=8) &65&421&4&0.31&12.06\\
			refuel (N=6, E=8) &270&1301&36&1.27&7.52\\
			refuel (N=7, E=7) &302&1545&35&0.97&4.55\\
			rocks (N=4) &331&3484&65&64.66&82.89\\
			\midrule
			\multicolumn{4}{l|}{\textbf{Total time}}&5643.25&351.81\\
			\bottomrule
		\end{tabular}
	}
\end{table*}

\cref{tab:time} reports the time taken for synthesizing the policy and creating the DT-FSCs from them.
While the model checker \storm\ can leverage parallelization, learning the DTs were done sequentially. 
Even then, creating DT-FSCs from synthesized policies takes less than 2 minutes per benchmark, even for large benchmarks, while policy synthesis can take over an hour for a large benchmark (see Table 5 in the Appendix for a detailed time comparison).
In total, \textbf{for the benchmarks in Set 1, the policy synthesis took 94 minutes, whereas DT-FSC creation was completed in less than 6 minutes}.
This highlights the efficiency of using DT-FSCs as a lightweight postprocessing step that adds explainability with minimal time overhead.
\FloatBarrier
\section{Reproducibility}
\label{app:reproducibility}
To ensure the reproducibility of our results, a docker image is publicly available  at the following link:
\begin{center}
	\url{https://doi.org/10.5281/zenodo.17304440}
\end{center}

\subsection{Experimental Environment}
The experiments were conducted on a host machine with the following specifications:
\begin{itemize}
	\item[\textbf{CPU:}] AMD Ryzen 7 PRO 6850U (16 cores, 2.7 GHz base clock)
	\item[\textbf{RAM:}] 32 GB
	\item[\textbf{Operating System:}] Ubuntu 24.04.1 LTS with Linux kernel 6.8.0-51
\end{itemize}
All experiments were executed within a Docker container using Docker version 26.1.3. The container was not restricted by any explicit resource limits, allowing it to fully utilize the resources of the host machine.

\subsection{Docker Artifact}
The following commands demonstrate how to use the Docker artifact:
\begin{verbatim}
	# Load the Docker image
	docker load < pomdp-dt-fsc.tar
	
	# Run the Docker container interactively
	docker run -it pomdp-dt-fsc:latest /bin/bash
	
	# Navigate to the repository
	cd pomdp-explainable-policy/
\end{verbatim}

The docker includes scripts to reproduce both benchmark set (1) (qualitative anaysis) and benchmark set (2) (quantitative analysis):
\begin{itemize}
	\item[\textbf{Qualitative Analysis:}]
	Navigate to the respective directory and run:
	\begin{verbatim}
		./run_all.sh
	\end{verbatim}
	\item[\textbf{Quantitative Analysis:}]
	Navigate to the quantitative directory and execute:
	\begin{verbatim}
		./run_quantitative.sh
	\end{verbatim}
\end{itemize}

\end{document}